\pdfoutput=1
\documentclass[10pt,twocolumn,letterpaper]{article}

\usepackage[pagenumbers]{cvpr} 

\usepackage[accsupp]{axessibility}

\usepackage{graphicx}
\usepackage{amsmath}
\usepackage{amssymb}
\usepackage{booktabs}
\usepackage{multirow}
\usepackage{makecell}
\usepackage[table]{xcolor}
\usepackage{colortbl}
\usepackage{graphicx}

\usepackage{enumitem}
\setenumerate[1]{itemsep=0pt,partopsep=0pt,parsep=\parskip,topsep=5pt}
\setitemize[1]{itemsep=0pt,partopsep=0pt,parsep=\parskip,topsep=5pt}
\setdescription{itemsep=0pt,partopsep=0pt,parsep=\parskip,topsep=5pt}

\usepackage[pagebackref,breaklinks,colorlinks,bookmarks=false]{hyperref}
\usepackage[capitalize]{cleveref}
\crefname{section}{Sec.}{Secs.}
\Crefname{section}{Section}{Sections}
\Crefname{table}{Table}{Tables}
\crefname{table}{Tab.}{Tabs.}

\def\cvprPaperID{4131} 
\def\confName{CVPR}
\def\confYear{2023}

\begin{document}

\title{NeuralPCI: Spatio-temporal Neural Field for 3D Point Cloud Multi-frame Non-linear Interpolation}

\author{Zehan Zheng$^{\ast}$,~ Danni Wu$^{\ast}$,~ Ruisi Lu,~ Fan Lu,~ Guang Chen$^{\dagger}$,~ Changjun Jiang\\
Tongji University\\
{\tt\small \{zhengzehan, woodannie, lrs910, lufan, guangchen, cjjiang\}@tongji.edu.cn}}
\maketitle

\renewcommand{\thefootnote}{}
\footnote{$^\ast$ Equal contribution. $^\dagger$ Corresponding author.}

\vspace{-0.5cm}

\begin{abstract}
\vspace{-0.1cm}
In recent years, there has been a significant increase in focus on the interpolation task of computer vision. Despite the tremendous advancement of video interpolation, point cloud interpolation remains insufficiently explored. Meanwhile, the existence of numerous nonlinear large motions in real-world scenarios makes the point cloud interpolation task more challenging. In light of these issues, we present \textbf{NeuralPCI}: an end-to-end 4D spatio-temporal \textbf{N}eural field for 3D \textbf{P}oint \textbf{C}loud \textbf{I}nterpolation, which implicitly integrates multi-frame information to handle nonlinear large motions for both indoor and outdoor scenarios. Furthermore, we construct a new multi-frame point cloud interpolation dataset called NL-Drive for large nonlinear motions in autonomous driving scenes to better demonstrate the superiority of our method. Ultimately, NeuralPCI achieves state-of-the-art performance on both DHB (Dynamic Human Bodies) and NL-Drive datasets. Beyond the interpolation task, our method can be naturally extended to point cloud extrapolation, morphing, and auto-labeling, which indicates its substantial potential in other domains. Codes are available at \url{https://github.com/ispc-lab/NeuralPCI}.  

\end{abstract}

\vspace{-0.1cm}

\section{Introduction}
\label{sec:intro}

In the field of computer vision, sequential point clouds are frequently utilized in many applications, such as VR/AR techniques~\cite{wang2021self, garrido2021point, zhang2021sequential} and autonomous driving~\cite{qi2021offboard, chen2021moving, yang2021auto4d}. The relatively low frequency of LiDAR compared to other sensors, i.e., 10–20 Hz, impedes exploration for high temporal resolution point clouds~\cite{zeng2022idea}. Therefore, interpolation tasks for point cloud sequences, which have not been substantially investigated, are receiving increasing attention.

\begin{figure}[t]
\centering
  \includegraphics[width=0.45\textwidth]{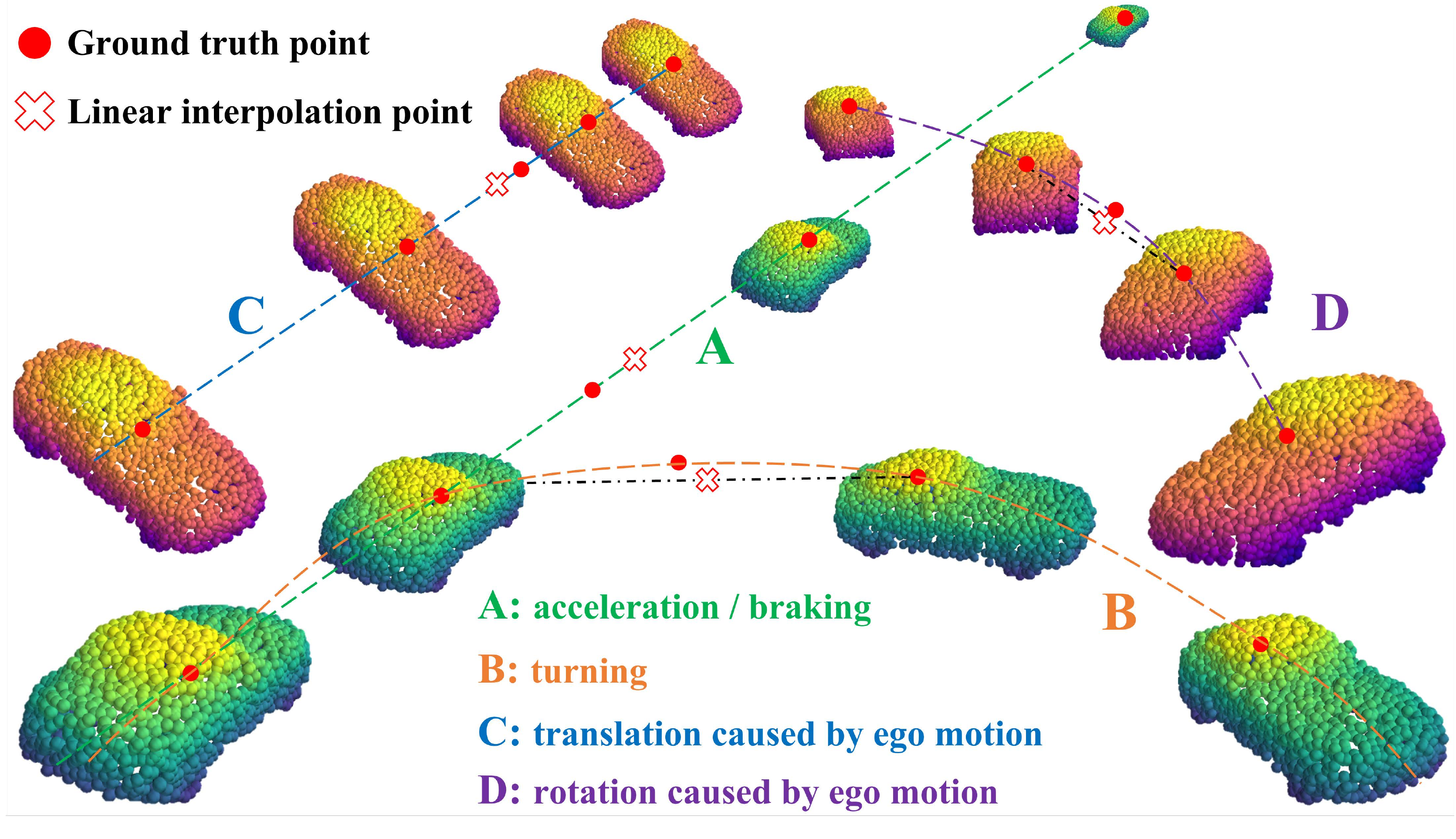}
\vspace{-0.2cm}
  \caption{\textbf{Common cases of nonlinear motions in autonomous driving scenarios.} Spatially uniform linear interpolation (\includegraphics[width=0.01\textwidth]{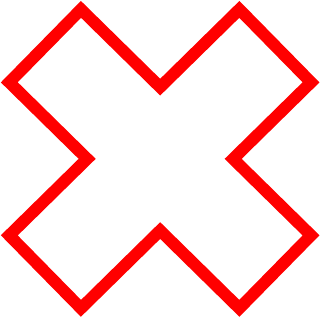}) using the middle two frames of the point cloud differs significantly from the actual situation (\includegraphics[width=0.009\textwidth]{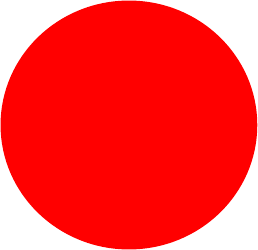}), so it is necessary to take multiple point clouds into consideration for nonlinear interpolation.}
  \label{fig:nonlinear scenes}
\vspace{-0.4cm}
\end{figure}  

With the similar goal of obtaining a smooth sequence with high temporal resolution, we can draw inspiration from the video frame interpolation (VFI) task. Several VFI methods~\cite{zhang2019multi, xu2019quadratic, liu2020enhanced, dutta2022non, shi2022video, chi2020all} concentrate on nonlinear movements in the real world. They take multiple frames as input and generate explicit multi-frame fusion results based on flow estimation~\cite{xu2019quadratic, zhang2019multi, liu2020enhanced, dutta2022non, chi2020all} or transformer~\cite{shi2022video}. Nonetheless, due to the unique structure of point clouds~\cite{qi2017pointnet}, it is non-trivial to extend VFI methods to the 3D domain.

Some early works~\cite{liu2020plin, liu2021pseudo} rely on stereo images to generate pseudo-LiDAR point cloud interpolation. For pure point cloud input, previous methods~\cite{lu2021pointinet, zeng2022idea} take two consecutive frames as input and output the point cloud at a given intermediate moment. However, with limited two input frames, these approaches can only produce linear interpolation results~\cite{lu2021pointinet}, or perform nonlinear compensation by fusing input frames in the feature dimension linearly~\cite{zeng2022idea}, which is inherently a data-driven approach to learning the dataset-specified distribution of nonlinear motions rather than an actual nonlinear interpolation. Only when the frame rate of the input point cloud sequence is high enough or the object motion is small enough, can the two adjacent point clouds satisfy the linear motion assumption. Nonetheless, there are numerous nonlinear motions in real-world cases. For instance, as illustrated in~\cref{fig:nonlinear scenes}, the result of linear interpolation between two adjacent point cloud frames has a large deviation from the actual situation. A point cloud sequence rather than just two point cloud frames allows us to view further into the past and future. Neighboring multiple point clouds contain additional spatial-temporal cues, namely different perspectives, complementary geometry, and implicit high-order motion information. Therefore, it is time to rethink the point cloud interpolation task with an expanded design space, which is an open challenge yet.

Methods that explicitly fuse multiple point cloud frames generally just approximate the motion model over time, which actually simplifies real-world complex motion. The neural field provides a more elegant way to parameterize the continuous point cloud sequence implicitly. Inspired by NeRF~\cite{mildenhall2021nerf} whose view synthesis of images is essentially an interpolation, we propose NeuralPCI, a neural field to exploit the spatial and temporal information of multi-frame point clouds. We build a 4D neural spatio-temporal field, which takes sequential 3D point clouds and the independent interpolation time as input, and predicts the in-between or future point cloud at the given time. Moreover, NeuralPCI is optimized on runtime in a self-supervised manner, without relying on costly ground truths, which makes it free from the out-of-the-distribution generalization problem. Our method can be flexibly applied to segmentation auto-labeling and morphing. Besides, we newly construct a challenging multi-frame point cloud interpolation dataset called NL-Drive from public autonomous driving datasets. Finally, we achieve state-of-the-art performance on indoor DHB dataset and outdoor NL-Drive dataset.

Our main contributions are summarized as follows:
\begin{itemize} 
\item We propose a novel multi-frame point cloud interpolation algorithm to deal with the nonlinear complex motion in real-world indoor and outdoor scenarios.

\item We introduce a 4D spatio-temporal neural field to integrate motion information implicitly over space and time to generate the in-between point cloud frames at the arbitrary given time.

\item A flexible unified framework to conduct both the interpolation and extrapolation, facilitating several applications as well. 
\end{itemize}

\section{Related Work}
\label{sec: related work}

\textbf{Video Frame Interpolation}. Most VFI approaches are based on optical flow estimation, focusing on the motion cues between two consecutive input frames. These methods warp source frames with the aid of the optical flow to generate the intermediate frame~\cite{jiang2018super, bao2019depth, reda2019unsupervised, park2020bmbc}. In order to deal with complex motions, some works exploit nonlinear information by expanding the design domain to multiple consecutive frames~\cite{xu2019quadratic, liu2020enhanced, dutta2022non, chi2020all, shi2022video}. QVI~\cite{xu2019quadratic} approximates the flow-based velocity and acceleration of the quadratic motion model explicitly. Following QVI, EQVI~\cite{liu2020enhanced} improves the training strategy. Recently, Dutta \etal~\cite{dutta2022non} utilizes space-time convolution to adaptively switch motion models with discontinuous motions. These works inspire us to design a point cloud interpolation network with multiple input frames. However, it is still challenging to extend these methods to the unordered and unstructured point cloud.

\textbf{Point Cloud Interpolation}. Existing point cloud interpolation methods try to find point-to-point correspondences between two point cloud reference frames. An intuitive way is to utilize scene flow, the extension of optical flow in the 3D domain. PointINet~\cite{lu2021pointinet} warps two input frames with bi-directional flows to the intermediate frame, then samples the two warped results adaptively and fuses them with attentive weights. This approach is performed under the linear motion assumption and relies heavily on the accuracy of the scene flow backbone. IDEA-Net~\cite{zeng2022idea} proposes an alternative method to solve the correlation between two input frames by learning a one-to-one alignment matrix and refining linear interpolation results with a trajectory compensation module. However, the one-to-one correspondence assumption limits its application for large-scale outdoor point cloud datasets. Moreover, the higher-order motion information in the temporal domain is overlooked~\cite{dutta2022non} when taking two frames as input. Therefore, there remain challenges to capturing and modeling the complex nonlinear motion in the real world. To address this issue, we propose a novel neural field that takes advantage of the more comprehensive spatio-temporal information of multiple point cloud frames.

\textbf{Neural Implicit Representation}. Distinct from the common paradigm of learning-based methods, neural fields are fitted to a single degraded sample rather than a large dataset of samples. The neural field can be seen as a parameterization of diverse input types, such as 2D images~\cite{chen2019net, karras2021alias}, 3D shapes~\cite{mescheder2019occupancy, park2019deepsdf}, \etc. Since NeRF~\cite{mildenhall2021nerf} presents a novel neural radiance field, which encodes a scene with spatial location and view direction as input and outputs volume rendering, several dynamic scene synthesis studies~\cite{du2021neural, gao2021dynamic, li2021neuralSF, park2021nerfies, pumarola2021d, tretschk2021non, xian2021space, li2021neural} have been proposed to exploit the representation ability of neural radiance fields for dynamic scenes. Based on the linear motion assumption, Li \etal~\cite{li2021neural} proposes a time-variant continuous neural representation for space-time view synthesis. NSFP~\cite{li2021neuralSF} presents a neural prior to regularize scene flow implicitly. Following NSFP, Wang \etal~\cite{wang2022neural} introduce a neural trajectory prior to representing the trajectories as a vector field. Coordinate-based networks show great potential for encoding a continuous input domain over arbitrary dimensions at an arbitrary resolution. Our work utilizes and exploits the ability of the coordinate-based network to represent continuous spatio-temporal motions of dynamic nonlinear scenes.

\begin{figure*}[ht]
\centering
  \includegraphics[width=0.9\textwidth]{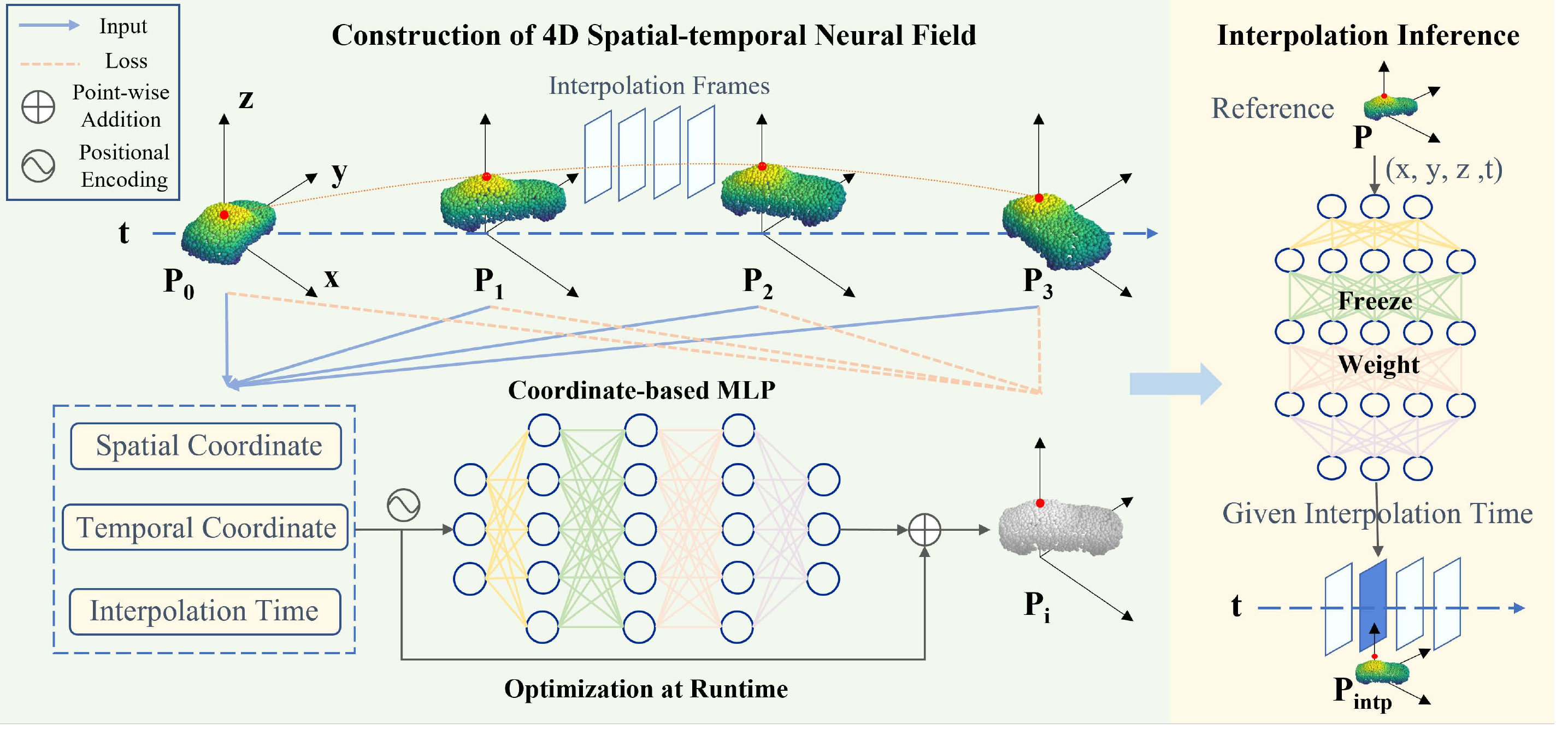}
  \caption{\textbf{Overview of our proposed NeuralPCI.} The 4D neural field is constructed by encoding the spatio-temporal coordinate of the multi-frame input point clouds via a coordinate-based multi-layer perceptron network. In the inference stage after self-supervised optimization, NeuralPCI receives a reference point cloud and an arbitrary interpolation frame moment as input to generate the point cloud of the associated spatio-temporal location.}
  \label{fig:overview}
\vspace{-.4cm}
\end{figure*}

\section{Methodology}
\label{sec: method}

In this section, we propose a novel end-to-end 4D spatio-temporal neural field for 3D point cloud interpolation named NeuralPCI. Firstly, we state the problem of multi-frame point cloud interpolation. ~\cref{sec: overview} then provides an overview of NeuralPCI's architecture and design philosophy. Following this, we explain the details of how to construct the 4D spatio-temporal neural field and integrate neighboring multi-frame nonlinear information of point clouds in ~\cref{sec: 4d spatio-temporal neural field,sec: multi-frame nonlinear integration}. Finally, in ~\cref{sec: self-supervised optimization}, we further elaborate on the self-supervised optimization manner of the whole neural field.

\textbf{Problem Formulation}.   
Let $P\in \mathbb{R}^{N\times3}$ be one frame of a dynamic point cloud sequence, with $N$ sampled points at the time $t\in \mathbb{R}$. Given a low temporal resolution sequence of $M$ frames of point clouds $S=\{P_{0}, P_{1}, ... ,P_{M-1}\}$ with their corresponding timestamps $T=\{t_{0}, t_{1}, ... ,t_{M-1}\}$, the goal of NeuralPCI is to predict the intermediate point cloud frame $P_{i}$ at an arbitrary given time $t_{i}$ for interpolation. Interpolating $n\in \mathbb{Z^{+}}$ frames of point clouds at equal intervals between every two consecutive frames yields a sequence of point clouds with $n+1$ times higher temporal resolution.
  
For the multi-frame point cloud interpolation task, we define the input as every four consecutive point cloud frames in the sequence $S$ and assume that these frames are equally time-spaced for convenience. The output is the temporally uniform interpolation result of $n$ point cloud frames between the middle two frames. Then, sliding the window of multi-frame inputs traverses the entire sequence.

\subsection{NeuralPCI Overview}
\label{sec: overview}

Mildenhall~\etal~\cite{mildenhall2021nerf} build the neural radiance field by inputting a series of 2D images with different viewpoints, and then generate the unknown image under a new viewpoint using neural volume rendering. Although the huge gap between images and point clouds makes it non-trivial to apply the neural field technique to the point cloud interpolation task, we still want to find a way to address the nonlinear interpolation problem of multi-frame point clouds under a similar design philosophy. That is, encoding a sequence of 3D point clouds at different moments to construct a 4D neural spatio-temporal field and then feeding an arbitrary interpolation frame moment into the network to generate the point cloud of the associated spatio-temporal location. 

Consequently, we propose NeuralPCI whose overall structure is depicted in \cref{fig:overview}. It can be divided into three main sections for elaboration. In the optimization stage, the 4D neural field is constructed by encoding the spatio-temporal coordinate of the multi-frame input point cloud via a coordinate-based Multi-Layer Perceptron (MLP) network. For each point cloud frame of the input, the interpolation time is set to the corresponding timestamps of four input frames in order to provide the network with the ability to generate the point cloud of the respective spatio-temporal position, and then optimize the neural field through self-supervised losses. In the inference stage, we run forward the neural field with the spatio-temporal coordinate of a reference point cloud and the moment of the interpolation frame as input to obtain the corresponding in-between point cloud.

\subsection{4D Spatio-temporal Neural Field}
\label{sec: 4d spatio-temporal neural field}

Following the statement in~\cite{xie2022neural}, a field is a quantity defined for all spatial and/or temporal coordinates, and a neural field is a field that is parameterized fully or in part by a neural network.
Here, we use a coordinate-based MLP network to represent the scenes, which takes as input the 3D spatial coordinate $\mathbf{x} \in \mathbb{R}^{3}$ and the 1D temporal coordinate $t \in \mathbb{R}$, and produces as output the motions $\Delta \mathbf{x} \in \mathbb{R}^{3}$. The MLP is parameterized by $\Theta$ and it can be viewed as a mapping from the coordinate field to the motion field.

Furthermore, we leverage an independent input, the interpolation frame time $t_{intp} \in \mathbb{R}$, to provide MLP with time cues and regulate the scene motion output at the interpolation moment relative to the original input spatio-temporal coordinates. Then, we conduct the point-wise addition of the interpolation frame motion and the input point cloud to generate the final output of the neural field, which is the 3D point cloud coordinate at the interpolation frame moment $\mathbf{x}_{intp} \in \mathbb{R}^{3}$. Formally, the neural field $\Phi$ is defined as:
\vspace{-.4cm}

\begin{equation}
\begin{split}
    \Phi \left(\mathbf{z} ; \Theta\right): ~\mathbf{x}_{intp} = \mathbf{x} + F_{\Theta}\left(\mathbf{z}\right)
\end{split}
\end{equation}

\noindent where $\mathbf{z}$ represents the spatial-temporal coordinate input of point clouds, and $F_{\Theta}$ is defined as:
\begin{equation}
\label{eq: MLP}
\begin{split}
    F_{\Theta}: \mathbb{R}^{5} \rightarrow \mathbb{R}^{3},\quad \Delta \mathbf{x} = F_{\Theta}\left(\mathbf{z}\right) = F_{\Theta}\left(\mathbf{x},t,t_{intp}\right)
\end{split}
\end{equation}

In such a manner, we implicitly construct a 4D, i.e., ($x,y,z,t$) in physical, spatio-temporal neural field to represent the entire scenarios of sequential point clouds. Eventually, we are able to utilize the continuity of the dedicated neural field to smoothly interpolate the point cloud at an arbitrary in-between moment.

\subsection{Multi-frame Nonlinear Integration}
\label{sec: multi-frame nonlinear integration}

In order to integrate multi-frame information, an intuitive way is utilizing the existing pair-frame point cloud interpolation algorithm for every two frames among the multi-frame point cloud input and fuse the predicted results. We adopt it as a baseline model (see \textit{Supplementary Material}), and it turns out that directly fusing multiple intermediate predictions by random sampling leads to worse results. 

Besides, an alternative way is to explicitly model the nonlinear kinematic equations taking advantage of multi-frame point clouds. We follow the nonlinear video interpolation algorithm~\cite{xu2019quadratic} and extend it to the 3D domain to formulate the high order equation of point clouds. Nonetheless, the explicit modeling approach is also ineffective in real-world scenarios with complex motions (see \textit{Supplementary Material}). Consequently, we integrate multi-frame point clouds more effectively using spatio-temporal neural fields.

Our proposed NeuralPCI does not impose a restriction on the frame number of the input point clouds, and thus can be naturally expanded to multi-frame point clouds. When the number of multi-frame inputs is 4, for example, the input set containing a point cloud sequence and corresponding timestamps are $S=\{P_{0}, P_{1}, P_{2}, P_{3}\}$ and $T=\{t_{0}, t_{1}, t_{2}, t_{3}\}$, respectively. As shown in \cref{fig:overview}, the neural field receives the point cloud $P_{0}$ and time $t_{0}$ as input, and the time step for interpolation $t_{intp}$ is adjusted to one of $\{t_{0}, t_{1}, t_{2}, t_{3}\}$ to yield the predicted point clouds $\{\hat{P}_{0}^{t_{0}}, \hat{P}_{0}^{t_{1}}, \hat{P}_{0}^{t_{2}}, \hat{P}_{0}^{t_{3}} \}$. The loss function (described in~\cref{sec: self-supervised optimization}) is then calculated between the prediction $\hat{P}_{i}^{t_{j}}$and each of the four input frames of point clouds $P_{j}$. The entire point cloud sequence is traversed through the same operation, with the spatio-temporal neural field end-to-end optimized meanwhile.

Each input frame serves as a constraint to supervise optimization, allowing us to incorporate the information from multiple frames of point clouds more elegantly. Owing to the continuity, smoothness, and excellent fitting ability of MLP, the derivative of the final spatio-temporal neural field function with respect to the interpolated frame time $t_{intp}$ is an implicit higher-order continuous function, which can better handle complex motion scenes and can yield smooth interpolation outputs.

\subsection{Self-supervised Optimization}  
\label{sec: self-supervised optimization}  
NeuralPCI optimizes the weights of the neural field in a self-supervised manner. As illustrated in ~\cref{fig:overview}, the interpolation time is adjusted as each timestamp of inputs to generate predictions for all input point cloud frames, and the neural field back-propagates the gradients to update the sample-specified parameters by minimizing the following distribution loss between the input and predicted point clouds.

\textbf{CD Loss}. Chamfer Distance (CD)~\cite{fan2017point} measures the distribution difference between two point clouds. We adopt CD as the main term in the loss function, which can be expressed as the following equation:  
\vspace{-.05cm}
\begin{small}
\begin{equation}
\setlength\belowdisplayskip{0.05cm}
\label{eq: CD}
\begin{split}
    \mathcal{L}_{CD}=\frac{1}{N} \sum_{\hat{p}_{i} \in \hat{P}} \min _{p_{i} \in P}\left\|\hat{p}_{i}-p_{i}\right\|^{2}_{2}
    +\frac{1}{N} \sum_{p_{i} \in P} \min _{\hat{p}_{i} \in \hat{P}}\left\|p_{i}-\hat{p}_{i}\right\|^{2}_{2}
\end{split}
\end{equation}
\end{small}
\vspace{-.05cm}
\noindent where $P$ and $\hat{P}$ are the input and predicted point cloud frames. $ p_{i} $ and $ \hat{p}_{i} $ represent the points in the respective point clouds. $ \| . \|_2$ denotes the $L_2$ norm of the spatial coordinate.

\textbf{EMD Loss}. Earth Mover's Distance (EMD)~\cite{rubner2000earth} calculates the corresponding points by solving the optimal transmission matrix of the two point clouds. We minimize the EMD loss to encourage the two point clouds to have the same density distribution, calculated as:
\vspace{-.1cm}
\begin{equation}
\label{eq: EMD}
     \mathcal{L}_{EMD}=\min _{\phi: \hat{P} \rightarrow P} \frac{1}{N} \sum_{\hat{p} \in \hat{P}}\|\hat{p}-\phi\left(\hat{p}\right)\|^{2}_{2}
\end{equation}
\vspace{-.4cm}

\noindent where $ \phi: \hat{P} \rightarrow P $ denotes a bijection from $ \hat{P} $ to $  P $. Due to its high computational complexity, we only use it in the loss function for sparse point clouds.

\textbf{Smoothness Loss}. To better regulate the estimated motion, we introduce the smoothness loss, which facilitates the interpolation of point clouds for local rigid motions and is utilized in autonomous driving scenarios.  
\vspace{-.05cm}
\begin{small}
\begin{equation}
\mathcal{L}_{S}=\sum_{p_{i} \in P} \frac{1}{\left|N\left(p_{i}\right)\right|} \sum_{p_{j} \in N\left(p_{i}\right)}\left\|\Delta \mathbf{x}_j -\Delta \mathbf{x}_i \right\|_{2}^{2}
\end{equation}
\end{small}

\noindent where $N(p_{i})$ stands for the nearest neighbors of the $i^{th}$ point $ p_{i} $ and $ \left| \cdot \right|$ denotes the number of points. And $ \Delta \mathbf{x}_i $ denotes the scene flow estimation  from point cloud $  P $ to $ \hat{P} $ (defined in~\cref{eq: MLP}) of the $i^{th}$ point $ p_{i} $.

\textbf{Total Loss Function}. We utilize the above three terms of loss and balance them with separately given weights (defined as $\alpha, \beta, \gamma$ in ~\cref{eq: alpha}) to obtain the total loss. We define multiple input frames and their corresponding timestamps as the input set. Then, we traverse the input set by assigning each point cloud of input frames as the reference and generate predictions of all input frames at their corresponding time steps. The overall loss is computed by summing the losses of every pair of point clouds as~\cref{eq: overall loss}.
\vspace{-0.1cm}
\begin{equation}
\setlength\belowdisplayskip{0.05cm}
\begin{split}
\label{eq: alpha}
    \Psi= \alpha \mathcal{L}_{CD} + \beta \mathcal{L}_{EMD} + \gamma \mathcal{L}_{S}
\end{split}
\end{equation}
\vspace{-.4cm}
\begin{equation}
\label{eq: overall loss}
\begin{split}
    \mathcal{L}=\sum_{P_i \in S} \sum_{t_j \in T} \Psi \left(P_{t_j}, \hat{P}_{i}^{t_{j}}\right)
\end{split}
\end{equation}

\noindent where $S$ and $T$ are the multiple input frames and their corresponding timestamps. $ P_{t_j} $ denotes the input frame at time $ t_j $ and $ \hat{P}_{i}^{t_{j}} $ denotes the neural field prediction when $ P_i $ is the reference frame and the interpolation time is $t_j$.

\section{Experiments}
\label{sec: experiments}
\subsection{Experimental Setup}
\label{sec: Experimental Setup}
\textbf{Datasets}. We evaluate NeuralPCI in both indoor and outdoor datasets. Indoor Dynamic Human Bodies dataset (DHB)~\cite{zeng2022idea} contains point cloud sequences for the non-rigid deformable 3D human motion with sampled 1024 points. We construct a challenging multi-frame interpolation dataset named Nonliner-Drive (NL-Drive) dataset. Based on the principle of hard-sample selection and the diversity of scenarios, NL-Drive dataset contains point cloud sequences with large nonlinear movements from three public large-scale autonomous driving datasets: KITTI~\cite{geiger2012we}, Argoverse~\cite{chang2019argoverse} and Nuscenes~\cite{caesar2020nuscenes}. More details of NL-Drive dataset are provided in \textit{Supplementary Material}.

\textbf{Baselines}. To demonstrate the performance of NeuralPCI, we compare our method with previous SOTA approaches, namely IDEA-Net~\cite{zeng2022idea} and PointINet~\cite{lu2021pointinet} for the interpolation task and MoNet~\cite{lu2021monet} for the extrapolation task. We reproduced the results on DHB Dataset of IDEA-Net and PointINet using official implementation. Moreover, we train PointINet and MoNet on NL-Drive dataset according to the official training setting (Since the training code of IDEA-Net has not been released yet, we do not report its results here). In addition, we utilize state-of-the-art scene flow estimation methods, i.e., neural-based NSFP~\cite{li2021neuralSF} and recurrent-based PV-RAFT~\cite{wei2021pv}, with linear interpolation to produce corresponding results on both DHB and NL-Drive datasets for comprehensive comparison. Remarkably, we evaluate all the optimization-based methods directly on the test set without using the training set.

\textbf{Metrics}. We adopt CD and EMD as quantitative evaluation metrics following~\cite{zeng2022idea,lu2021pointinet,lu2021monet}, which are described in~\cref{eq: CD,eq: EMD}, respectively.

\subsection{Implementation Details}

We implement NeuralPCI with PyTorch~\cite{paszke2019pytorch}. We define our NeuralPCI as a coordinate-based 8-layer MLP architecture with 512 units per layer and adopt LeakyReLU as the activation function. The network weights are randomly initialized with the Adam~\cite{kingma2014adam} optimizer (the $lr$ is 0.001). For each sample, the maximum optimization step is limited to 1000 iterations. The spatio-temporal coordinates of the point clouds are position-encoded by a sinusoidal function and fed into the MLP, while the interpolation time is inserted into the penultimate hidden layer to control the final output. All experiments were conducted on a single NVIDIA GeForce RTX 3090 GPU. For more implementation details, please refer to \textit{Supplementary Material}.

\begin{table*}[htbp]
\renewcommand\arraystretch{1.1}
\begin{minipage}[ht]{\textwidth}
\caption{\textbf{Quantitative comparison $\left( \times 10^{-3} \right)$ with other open-sourced methods on DHB-Dataset~\cite{zeng2022idea}.} Baseline methods include IDEA-Net~\cite{zeng2022idea}, PointINet~\cite{lu2021pointinet} and the results of linear interpolation of scene flow estimated by NSFP~\cite{li2021neural} and PV-RAFT~\cite{wei2021pv}.}
\centering
\label{tab: DHB-Dataset results}

\centering
\scalebox{0.95}{
\begin{tabular}{p{0.1\textwidth}<{\centering}cccccccccccccc}
\hline
\multirow{2}{*}{Methods} & 
\multicolumn{2}{c}{Longdress} & 
\multicolumn{2}{c}{Loot} & 
\multicolumn{2}{c}{Red\&Black} & 
\multicolumn{2}{c}{Soldier} & 
\multicolumn{2}{c}{Squat} & 
\multicolumn{2}{c}{Swing} & 
\multicolumn{2}{c}{Overall} \\ 
\cline{2-15}
& CD & EMD & CD & EMD & CD & EMD & CD & EMD & CD & EMD & CD & EMD & CD~$\downarrow$ & EMD~$\downarrow$ \\ 
\hline
IDEA-Net                                           
    & 0.89 & 6.01 & 0.86 & 8.62 & 0.94 & 10.34 & 1.63 & 30.07 & 0.62 & 6.68 & 1.24 & 6.93 & 1.02 & 12.03 \\
PointINet
    & 0.98 & 10.87 & 0.85 & 12.10 & 0.87 & 10.68 & 0.97 & 12.39 & 0.90 & 13.99 & 1.45 & 14.81 & 0.96 & 12.25  \\
NSFP                                             
    & 1.04 & 7.45 & 0.81 & 7.13 & 0.97 & 8.14 & 0.68 & 5.25 & 1.14 & 7.97 & 3.09 & 11.39 & 1.22 & 7.81 \\
PV-RAFT                                
    & 1.03 & 6.88 & 0.82 & 5.99 & 0.94 & 7.03 & 0.91 & 5.31 & 0.57 & 2.81 & 1.42 & 10.54 & 0.92 & 6.14 \\
\rowcolor{lightgray}  \makecell{ NeuralPCI }
    & \textbf{0.70} & \textbf{4.36} & \textbf{0.61} & \textbf{4.76} & \textbf{0.67} & \textbf{4.79} & \textbf{0.59} & \textbf{4.63} & \textbf{0.03} & \textbf{0.02} & \textbf{0.53} & \textbf{2.22} & \textbf{0.54} & \textbf{3.68}   \\
\hline
\end{tabular}}
\end{minipage}
\end{table*}





\begin{table*}
\renewcommand\arraystretch{1.05}
\begin{minipage}[ht]{\textwidth}
\caption{\textbf{Quantitative comparison with other open-sourced methods on NL-Drive Dataset.} \textit{Type} indicates whether the interpolation results are based on forward, backward, bidirectional flow, or neural field. \textit{Frame-1}, \textit{Frame-2} and \textit{Frame-3} refer to the three interpolation frames located at equal intervals between the two intermediate input frames.}
\centering
\label{tab: Nonlinear-Dataset results}

\centering
\scalebox{1}{
\begin{tabular}{p{0.14\textwidth}<{\centering}p{0.15\textwidth}<{\centering}p{0.03\textwidth}<{\centering}p{0.08\textwidth}<{\centering}p{0.03\textwidth}<{\centering}p{0.08\textwidth}<{\centering}p{0.03\textwidth}<{\centering}p{0.08\textwidth}<{\centering}p{0.05\textwidth}<{\centering}p{0.08\textwidth}<{\centering}}
\hline
\multirow{2}{*}{Methods} & 
\multirow{2}{*}{Type} & 
\multicolumn{2}{c}{Frame-1} & 
\multicolumn{2}{c}{Frame-2} & 
\multicolumn{2}{c}{Frame-3} & 
\multicolumn{2}{c}{Average} \\ 
\cline{3-10}
&   & CD & EMD & CD & EMD & CD & EMD & CD~$\downarrow$ & EMD~$\downarrow$ \\          
\hline
\multirow{2}{*}{NSFP}                                        
    & forward flow & 0.94 & 95.18  & 1.75 & 132.30 & 2.55 & 168.91 & 1.75 & 132.13  \\  
    & backward flow & 2.53 & 168.75 & 1.74 & 132.19 & 0.95 & 95.23  & 1.74 & 132.05  \\  
\multirow{2}{*}{PV-RAFT}
    & forward flow & 1.36 & 104.57 & 1.92 & 146.87 & 1.63 & 169.82 & 1.64 & 140.42  \\
    & backward flow & 1.58 & 173.18 & 1.85 & 145.48 & 1.30 & 102.71 & 1.58 & 140.46  \\
PointINet                                           
    & bi-directional flow & 0.93 & 97.48  & 1.24 & \textbf{110.22} & 1.01 & 95.65  & 1.06 & 101.12   \\ 
\rowcolor{lightgray} NeuralPCI                         
    & neural field & \textbf{0.72} & \textbf{89.03}  & \textbf{0.94} & 113.45 & \textbf{0.74} & \textbf{88.61}  & \textbf{0.80} & \textbf{97.03}  \\  

\hline
\end{tabular}}
\end{minipage}
\vspace{-.4ex}
\end{table*}




\subsection{Evaluation of Point Cloud Interpolation}
\begin{figure}[ht]
\centering
  \includegraphics[width=0.48\textwidth]{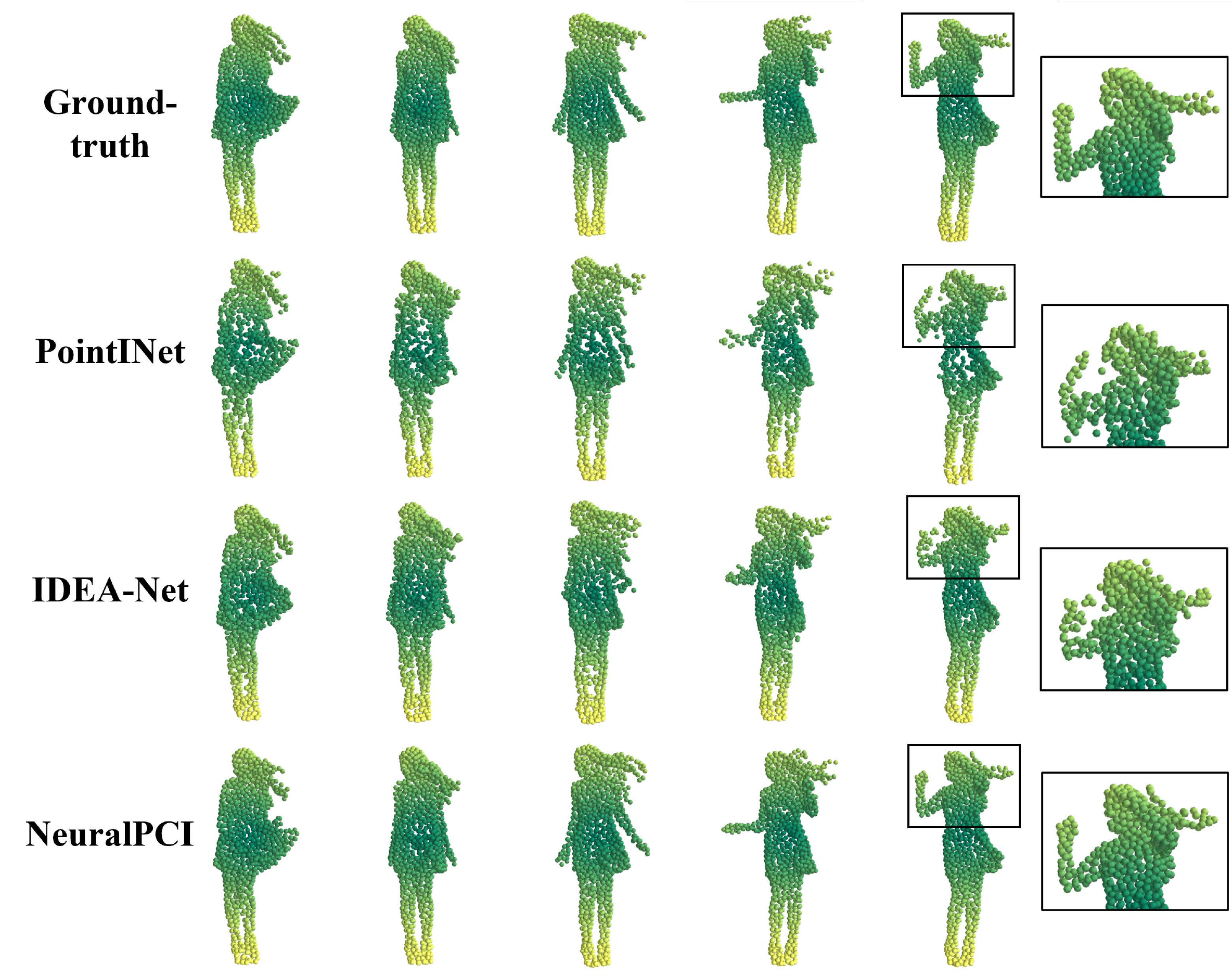}
  \caption{\textbf{Qualitative results on DHB dataset.} Each column represents one interpolation frame result in the point cloud sequence, where our method is more consistent with the ground truth and preserves local details better than previous SOTA methods.}
  \label{fig:DHB_vis}
\vspace{-.4cm}
\end{figure}

\textbf{Results on DHB dataset}. The quantitative comparison on DHB dataset is displayed in~\Cref{tab: DHB-Dataset results}, where our NeuralPCI outperforms other baseline approaches by a large margin. In particular, our overall CD is about half that of other baselines, and our overall EMD is \textbf{40\%} lower than the suboptimal PV-RAFT~\cite{wei2021pv}. As can be seen, IDEA-Net~\cite{zeng2022idea} does not perform well in all scenarios, but in contrast, our method achieves the best results in every single scene, especially in the \textit{Squat} scenario. This demonstrates the flexibility and adaptability of the neural field in various indoor scenarios. Furthermore, \cref{fig:DHB_vis} clearly exhibits the outcomes of qualitative experiments. The interpolation results of PointINet~\cite{lu2021pointinet} and IDEA-Net contain several outlier noise points and lack lots of detailed information in local areas, such as hair, hands, and skirt hems. Instead, our method benefits from the higher-order implicit function of the neural field, which better handles these complex motions.

\begin{figure*}[t]
\centering
  \includegraphics[width=0.88\textwidth]{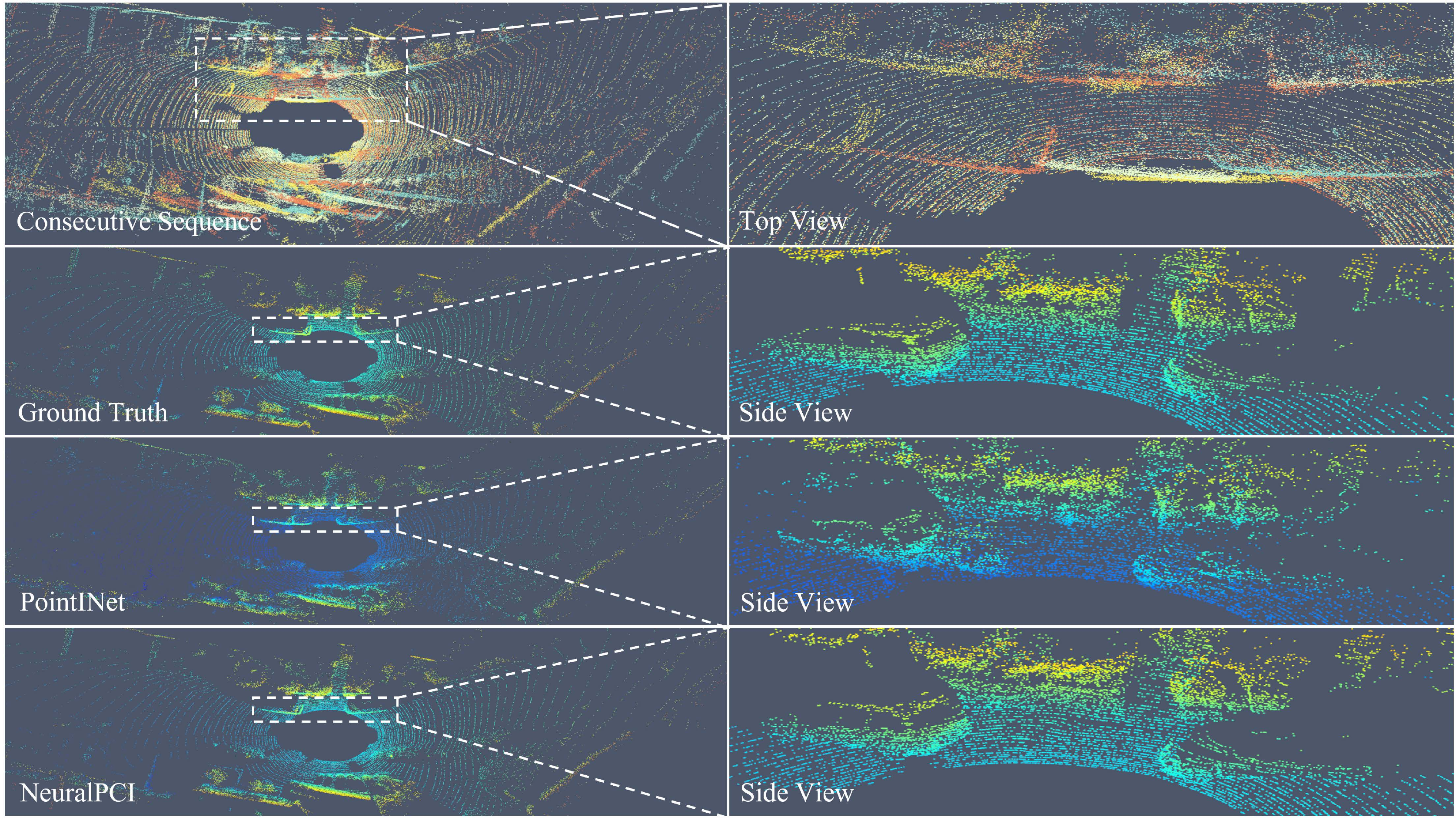}
  \caption{\textbf{Qualitative results on NL-Drive dataset.} We transform four frames of the input point cloud to the same coordinate system, and the overall motion of the point cloud sequence is depicted in the first row. The following three rows compare the interpolation results, demonstrating that our approach is more accurate and robust, whereas interpolation results of PointINet~\cite{lu2021pointinet} are noisier.}
  \label{fig:NL_vis}
\vspace{-.4cm}
\end{figure*}

\textbf{Results on NL-Drive dataset}. \Cref{tab: Nonlinear-Dataset results} shows results on NL-Drive dataset, where NeuralPCI achieves the best results on most frames and overall metrics. For instance, our method reaches comparable EMD error in the intermediate frames and significantly reduces the CD error, eventually outperforming SOTA by \textbf{24.5\%} and 4\% in the overall results of CD and EMD metrics, respectively. In the qualitative experiments, we show in detail the nonlinear motion in the outdoor autonomous driving scenario as well as the interpolation frame comparison results in~\cref{fig:NL_vis}. This indicates that NeuralPCI is scalable to large dense point clouds. As noted in the local zoomed-in view, the vehicle edges are clearer and sharper in the results of our method, while they are blurrier and noisier in the results of PointINet~\cite{lu2021pointinet}.

\begin{figure}[t]
\centering
  \includegraphics[width=0.48\textwidth]{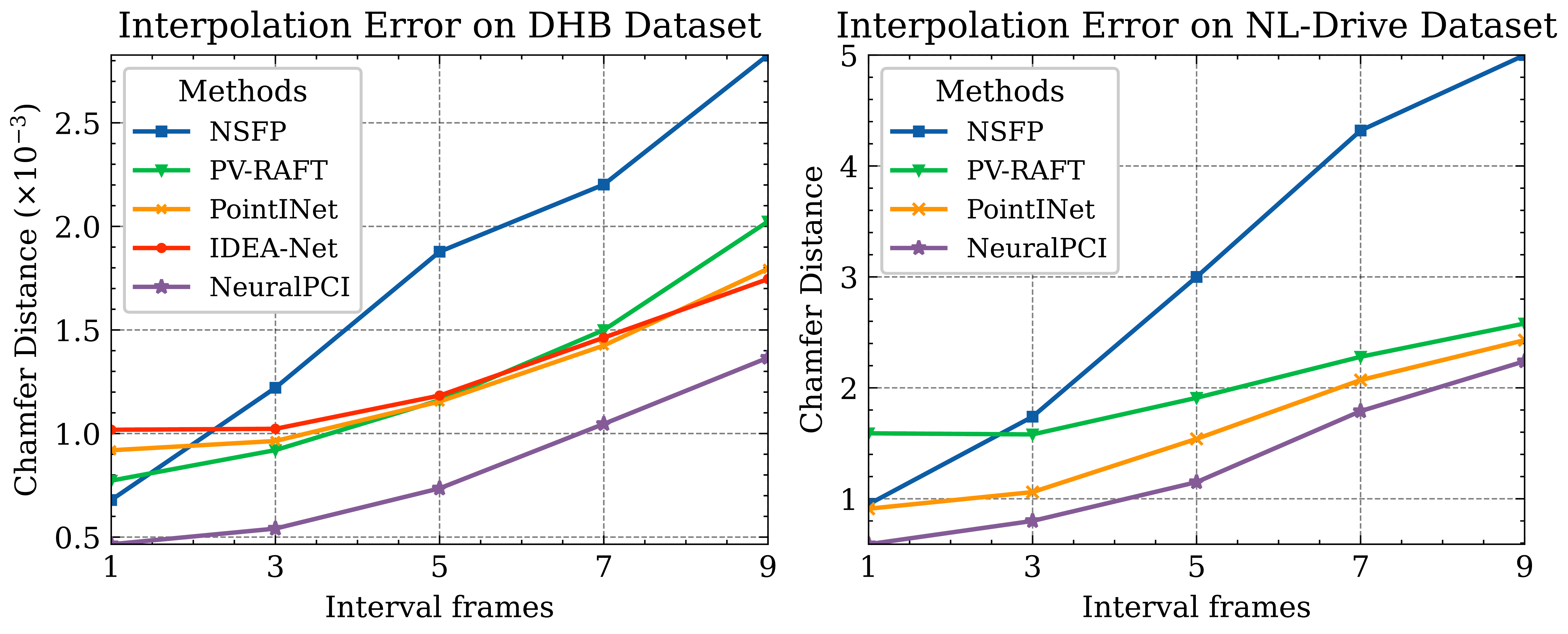}
  \caption{\textbf{Quantitative experiments at different intervals.} The CD error increases as the input frame interval grows. Our method is always optimal at different time intervals and has good robustness under long-distance motion.}
  \label{fig:intervals}
\vspace{-.4cm}
\end{figure}

\textbf{Additional Evaluation. }To evaluate the capability of NeuralPCI under large motions, we conduct additional experiments to observe the robustness of each method with increasing time interval between input frames. As illustrated in~\cref{fig:intervals}, the results on both datasets show that the performance of our method still lies at the optimum under large motion interpolation.

\subsection{Evaluation of Point Cloud Extrapolation}
As a parameterization of continuous point clouds over space and time, NeuralPCI is generalizable to predict the near future frame by adjusting the timestamp, making it more flexible than existing methods. We conduct extrapolation experiments on NL-Drive dataset with the same input settings as the interpolation task, while the outputs are four consecutive point cloud frames in the future. According to recent works about point cloud prediction~\cite{weng2021inverting,mersch2022self,weng2022s2net,lu2021monet}, we adopt the state-of-the-art method MoNet~\cite{lu2021monet} and also the linear extrapolation results based on scene flow from PV-RAFT~\cite{wei2021pv} and NSFP~\cite{li2021neuralSF} as baseline methods.



\begin{table}[t]
\renewcommand\arraystretch{1.5}
\centering
\caption{\textbf{Quantitative comparison with other baseline methods for point cloud extrapolation on NL-Drive dataset.} Here we adopt MoNet~\cite{lu2021monet} and linear extrapolation results based on scene flow from PV-RAFT~\cite{wei2021pv} and NSFP~\cite{li2021neuralSF}. \textit{Frames 1-4} refer to the consecutive extrapolation frames after the last input frame.}
\label{tab: pred}
\scalebox{0.75}{
\begin{tabular}{p{0.085\textwidth}<{\centering}p{0.03\textwidth}<{\centering}p{0.05\textwidth}<{\centering}p{0.03\textwidth}<{\centering}p{0.05\textwidth}<{\centering}p{0.03\textwidth}<{\centering}p{0.05\textwidth}<{\centering}p{0.04\textwidth}<{\centering}p{0.055\textwidth}<{\centering}}
\hline
\multirow{2}{*}{Methods} & \multicolumn{2}{c}{\makecell{Frame 1}} & \multicolumn{2}{c}{\makecell{Frame 2}} & \multicolumn{2}{c}{\makecell{Frame 3}} & \multicolumn{2}{c}{\makecell{Frame 4}}  \\ 
\cline{2-9}
& CD & EMD & CD & EMD & CD & EMD & CD$\downarrow$ & EMD$\downarrow$ \\ 
\hline
NSFP & 4.70 & 209.96 & 5.45 & 233.33 & 6.24 & 254.72 & 6.61  & 268.44 \\
PV-RAFT & 2.05 & 206.93 & 3.90 & 248.94 & 6.55 & 293.27 & 10.02 & 377.25 \\
MoNet & \textbf{0.66} & \textbf{81.90}  & \textbf{0.96} & \textbf{108.41} & \textbf{1.28} & \underline{135.96}  & \textbf{1.37} & 159.20 \\
\rowcolor{lightgray} NeuralPCI & \underline{0.78} & \underline{84.26} & \underline{1.20} & \underline{108.43}  & \underline{1.61} & \textbf{135.42} & \underline{1.87} & \textbf{156.64} \\
\hline
\end{tabular}}
\vspace{-.4cm}
\end{table}

 A quantitative comparison with baseline methods is shown in \Cref{tab: pred}. Flow-based methods suffer a sharp growth in error when the time for the future frame increases. Our method surpasses MoNet in terms of EMD error in the last two frames and achieves suboptimal results in the other remaining frames. MoNet's RNN module predicts each frame recurrently, while our neural field lacks constraints on the future direction. Despite that, NeuralPCI still achieves comparable results, indicating its flexibility to accomplish both inter-/extra-polation in a unified framework.

\subsection{Ablation Study}
\label{sec: ablation study}
Contributions of key modules in NeuralPCI, namely the multi-frame integration, NN-intp, EMD loss, smoothness loss, and network enhancement are shown in \Cref{tab: ablation}.

\begin{table}[t]
\renewcommand\arraystretch{1.1}
\centering
\caption{\textbf{Quantitative results  of ablation studies on DHB dataset~\cite{zeng2022idea} ($\times 10 ^{-3}$) and NL-Drive dataset.} Methods A$\sim$F denote CD loss, multi-frame integration, NN-intp, EMD loss, smoothness loss and network enhancement, respectively.}
\label{tab: ablation}
\scalebox{0.85}{
\begin{tabular}{cccccccccc} 
\hline
\multirow{2}{*}{\textbf{Datasets }} & \multirow{2}{*}{\textbf{ID }} & \multicolumn{6}{c}{\textbf{Methods }} & \multicolumn{2}{c}{\textbf{ Metrics }}  \\
\cline{3-10}
& & A & B & C & D & E & F & CD $\downarrow$ & EMD $\downarrow$ \\ 
\hline
\multirow{5}{*}{DHB}               
& 1 & \checkmark &  &  &  &  &  & 0.65 & 5.24 \\
& 2 & \checkmark & \checkmark &  &   &  &  & 0.59 & 5.08 \\
& 3 & \checkmark & \checkmark & \checkmark &  &  &  & 0.57 & 4.28 \\
& 4 & \checkmark & \checkmark & \checkmark & \checkmark  &  &  & 0.56 & 3.84 \\
& 5 & \checkmark & \checkmark & \checkmark & \checkmark &  & \checkmark & \textbf{0.54} & \textbf{3.68} \\
\hline
\multirow{5}{*}{NL-Drive}        
& 6 &  \checkmark &  &   &  &   &  & 0.86 & 114.31 \\
& 7 & \checkmark & \checkmark &  &  &  &  & 0.84 & 112.03 \\
& 8 & \checkmark & \checkmark & \checkmark &  &  &  & 0.81 & 104.71 \\
& 9 & \checkmark & \checkmark & \checkmark &  & \checkmark &  & 0.82 & 99.25 \\
& 10 & \checkmark & \checkmark & \checkmark  &  & \checkmark & \checkmark  & \textbf{0.80} & \textbf{97.03} \\
\hline
\end{tabular}}
\vspace{-.4cm}
\end{table}

\begin{figure*}[ht]
\centering
  \includegraphics[width=0.9\textwidth]{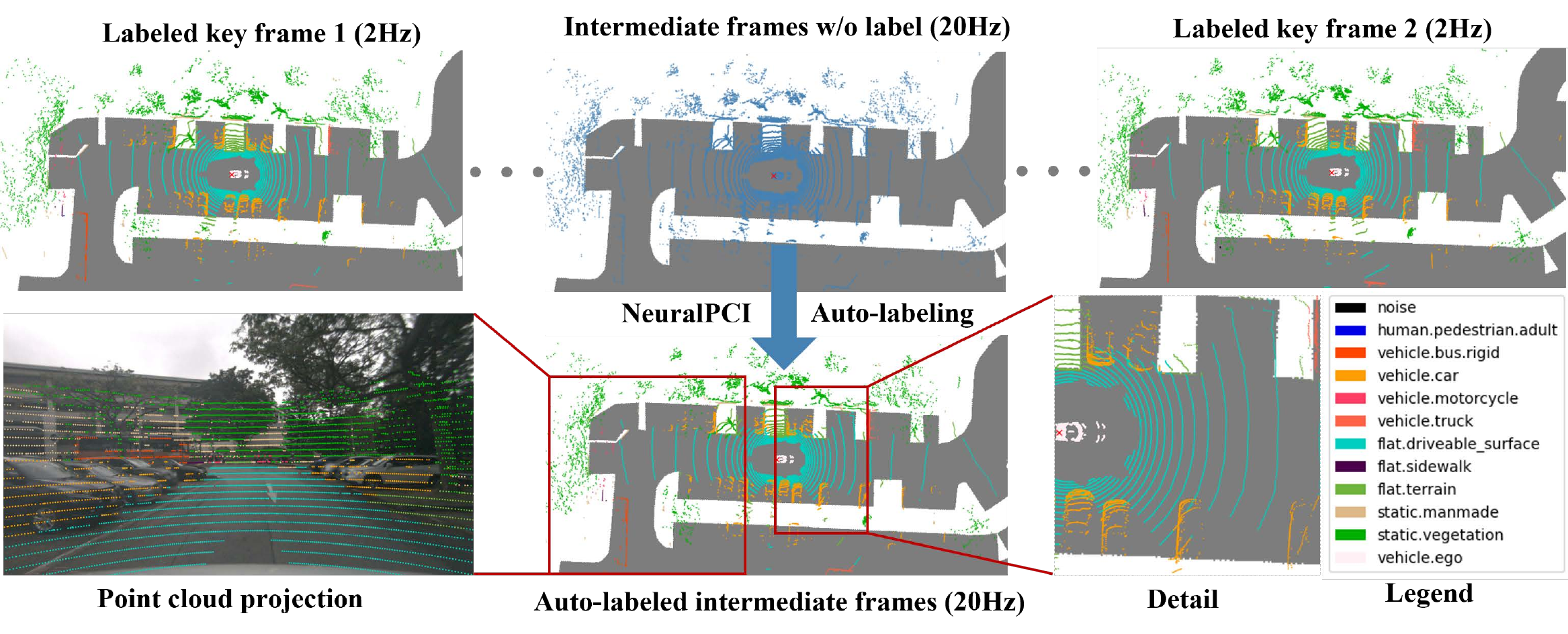}
  \vspace{-.2cm}
  \caption{\textbf{Visual results of NeuralPCI based auto-labeling.} We use NeuralPCI to take the labeled keyframe point clouds as input, output the interpolation results, and automatically assign labels to the intermediate frames. The second row shows the results of the auto-labeling, which intuitively achieves high labeling accuracy.}
  \label{fig:auto-label}
\vspace{-.4cm}
\end{figure*}  

\begin{figure}[ht]
\centering
  \includegraphics[width=0.47\textwidth]{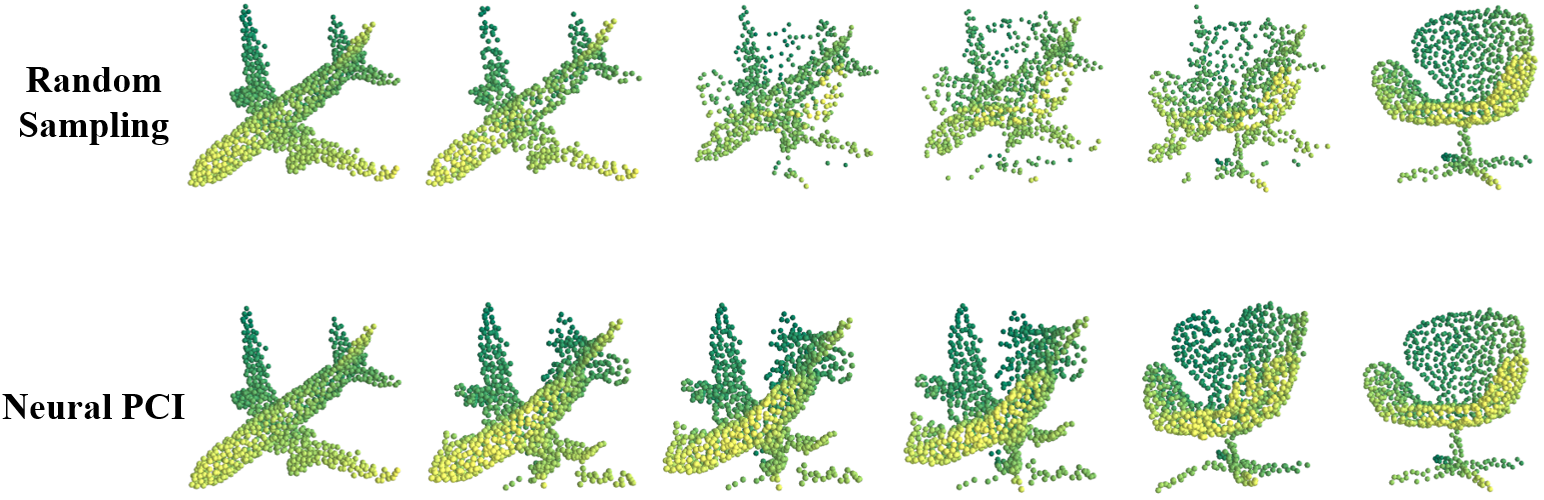}
  \caption{\textbf{The process of point clouds morphing} between the \textit{airplane} and the \textit{chair} samples in ModelNet40~\cite{wu20153d} using NeuralPCI.}
  \label{fig:morphing}
\vspace{-.4cm}
\end{figure}  

\textbf{Multi-frame and NN-intp.} We begin with a simple neural field with pair-frame input and optimized using only CD loss. With the multi-frame integration (ID 2\&7), NeuralPCI gains \textbf{9.2\%} and \textbf{3.1\%} reductions in CD and EMD error on DHB dataset, and \textbf{2.3\%} and \textbf{2.0\%} on NL-Drive dataset. By adopting the nearest neighbor in the time domain as the reference frame (ID 3\&8), the long-term error growth is effectively reduced, and another EMD improvement of \textbf{15.7\%} and \textbf{6.5\%} is achieved on the respective datasets.

\textbf{EMD and smoothness loss.} To prevent the network from overfitting on a single CD metric, we introduce additional loss terms to regulate the output. The extra EMD loss and smoothness loss help NeuralPCI achieve a significant reduction in EMD metric error, namely \textbf{10.3\%} (ID 4) on DHB dataset and \textbf{5.2\%} (ID 9) on NL-Drive dataset.

\textbf{Network enhancement.} The impact of network structure is future investigated. Previously, NeuralPCI utilized an 8-layer 256-hidden-unit MLP with a ReLU activation function and received the direct spatio-temporal coordinate as input. Here, we introduce a sinusoidal function-based position encoding, switch to the LeakyReLU activation function, and increase the width of the MLP. In the end, the network enhancement brings another improvement of \textbf{3.6\%} and \textbf{4.2\%} (ID 5) on DHB dataset and \textbf{2.4\%} and \textbf{2.2\%} (ID 10) on NL-Drive dataset.

\subsection{Applications}

\textbf{Auto-labeling. }The annotated point clouds with per-point segmentation GT in Nuscenes~\cite{caesar2020nuscenes} are at 2 Hz, only one tenth of the dataset. Even so, the labeling workload was laborious and enormous with a total of 1,400,000,000 points. Here, we explore the potential of NeuralPCI to generate keyframe-based interpolation results and assign point-wise labels to unannotated intermediate frames, which shows remarkable capability as depicted in~\cref{fig:auto-label}. We use four keyframes as input to predict intermediate point clouds, so the predicted outputs could inherit the corresponding keyframe labels in order. Then the $\mathit{k}$NN algorithm is applied between the unlabeled intermediate frame and the labeled predicted frame to annotate each ground-truth point.

\textbf{Point cloud morphing.} In addition to the conventional meaning of nonlinear motion under indoor and outdoor scenes, the transformation relationship across point clouds with totally different topological shapes is also needed. This transformation, i.e., point cloud morphing, is of interest for computer graphics simulation and data enhancement. We deploy NeuralPCI to output interpolation point clouds between two different classes of object point clouds under the ModelNet40~\cite{wu20153d} dataset to establish the process of point cloud morphing. As illustrated in~\cref{fig:morphing}, our method enables a more natural and smooth transformation between different shapes compared to random sampling.

\section{Conclusion}
\label{sec: conclusion}

In this paper, we redefine the input domain of the point cloud interpolation task as multiple consecutive frames instead of the two consecutive frames used in previous works, which increases the receptive field of the time domain. To achieve that, we presented NeuralPCI, a 4D spatio-temporal neural field for 3D point cloud interpolation that is able to implicitly integrate multi-frame information to handle nonlinear large motions. Our approach achieves state-of-the-art results in both indoor and outdoor datasets. Since neural-field-based methods are optimized at runtime, the application of our method is limited in terms of real-time efficiency. Based on NeuralPCI, further development can be considered in the future by improving its real-time performance and generalization over unknown samples.

\vspace{0.1cm}
\noindent
\textbf{Acknowledgments\quad}This work is supported by Shanghai Municipal Science and Technology Major Project (No.2018SHZDZX01), ZJ Lab, and Shanghai Center for Brain Science and Brain-Inspired Technology, and the Shanghai Rising Star Program (No.21QC1400900) and Tongji-Westwell Autonomous Vehicle Joint Lab Project.
\noindent


\newlength{\bibitemsep}\setlength{\bibitemsep}{.1\baselineskip}
\newlength{\bibparskip}\setlength{\bibparskip}{0pt}
\let\oldthebibliography\thebibliography
\renewcommand\thebibliography[1]{%
  \oldthebibliography{#1}%
  \setlength{\parskip}{\bibitemsep}%
  \setlength{\itemsep}{\bibparskip}%
}

{\small
\bibliographystyle{ieee_fullname}
\bibliography{NeuralPCI}
}

\noindent{\Large \textbf{Appendix}}

\appendix
\renewcommand{\appendixname}{\appendixname~\Alph{section}}
\renewcommand*{\thefigure}{S\arabic{figure}}
\renewcommand*{\thetable}{S\arabic{table}}

\section{Overview}
In this document, we present more details and several extra results as well as visualization. In~\cref{sec:appendix dateset details}, we introduce details of the datasets used in our work. Then we elaborate on the implementation details of our NeuralPCI and other baselines in~\cref{sec:appendix method details}. And in~\cref{sec:appendix experiments}, we provide extra results in multiple aspects, such as the convergence, different numbers of input frames, explicit versus implicit frame interpolation, varying point cloud densities and ground point removal. Finally, we show more qualitative results in~\cref{sec:appendix qualitative experiments}.


\section{Dataset Details}
\label{sec:appendix dateset details}
In this section, we introduce DHB dataset and the open-source autonomous driving datasets based Non-Linear Drive (NL-Drive) dataset. The train/val/test split of datasets facilitates the comparison of benchmarks. NeuralPCI optimizes at run-time, so it doesn’t need the training data. We perform NeuralPCI as well as NSFP~\cite{li2021neuralSF} on the test set directly to obtain the evaluation results. Other learning-based methods are pre-trained on the training set first and then compared on the same test set.

\subsection{DHB Dataset}

DHB dataset~\cite{zeng2022idea} consists of 14 point cloud sequences indicating dynamic human bodies, in which each point cloud frame is sampled to 1024 points. To align with baseline method ~\cite{zeng2022idea}, we adopt six sequence with 1,600 frames, (i.e., \textit{Longdress, Loot, Redandblack, Soldier, Squat\_2, Swing}) as the test dataset, and the remaining eight sequences with 1,600 frames as the train dataset. 

\begin{figure}[ht]
\centering
  \includegraphics[width=0.475\textwidth]{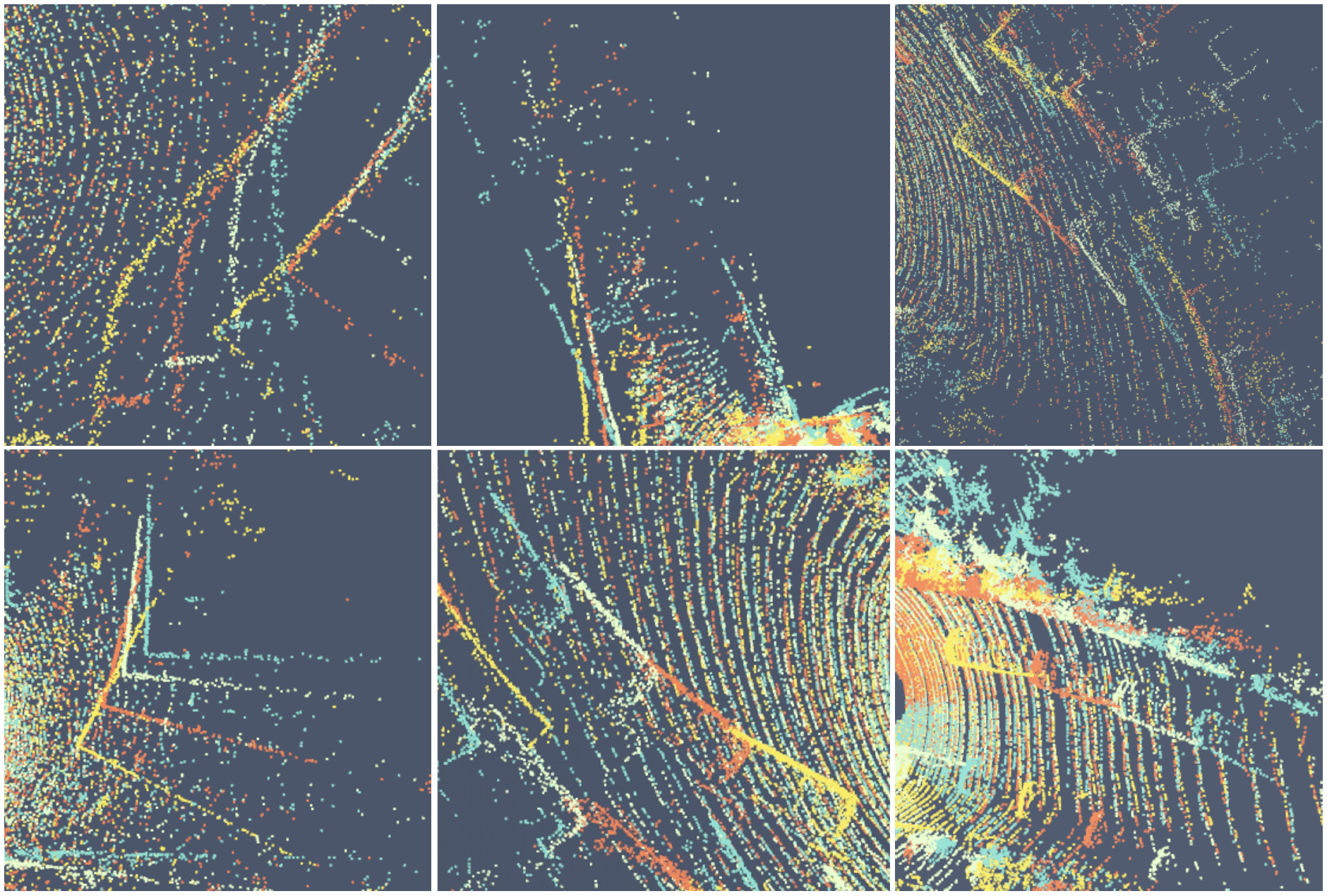}
  \caption{\textbf{Visualization (zoom-in view) for the point clouds of four consecutive input frames with equal time interval in the NL-Drive dataset.} The 1st to 4th frames sorted by chronological order are colored in blue, white, red and yellow, respectively. The motion of surrounding objects is nonlinear.}
  \label{fig:sample}
\vspace{-.4cm}
\end{figure}

\begin{figure*}[ht]
\centering
  \includegraphics[width=0.9\textwidth]{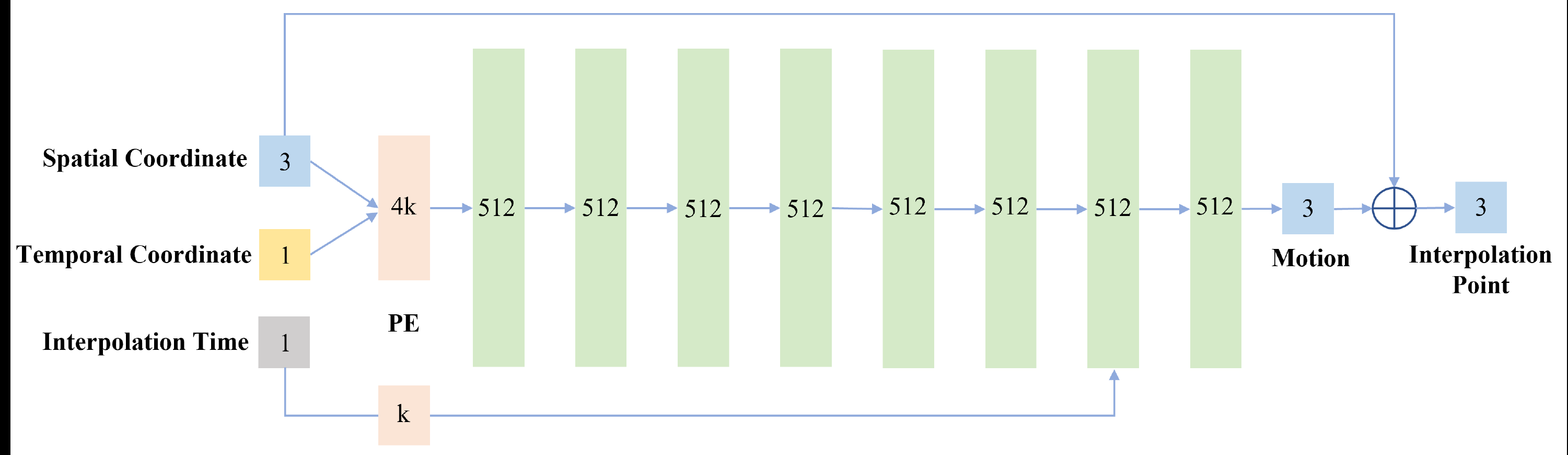}
  \caption{\textbf{Network architecture of our proposed NeuralPCI.} }
  \label{fig:network}
\vspace{-.4cm}
\end{figure*}

\subsection{NL-Drive Dataset}

We construct NL-Drive dataset based on three public autonomous driving datasets, namely KITTI odometry dataset~\cite{geiger2012we}, Argoverse 2 sensor dataset~\cite{chang2019argoverse}, and Nuscenes dataset~\cite{caesar2020nuscenes}. KITTI odometry dataset contains 22 LiDAR point cloud sequences in total, 11 sequences with ground truth (00$\sim$10), and we use 00$\sim$06 for training, 07$\sim$08 for validation, and the others for test. Argoverse 2 sensor dataset is composed of 1,000 scenarios with 150 LiDAR sweeps per scenario on average, while Nuscenes dataset consists of 1,000 driving scenes with about 400 LiDAR frames for each scene. For both datasets, we utilize the top 700 scenes to train, 701$\sim$850 scenes to validate and the remaining 150 scenes to test. Thus, we define the data source of the NL-Drive dataset as the mentioned splited datasets based on the training, validation and test ratio of 14:3:3. For Nuscenes dataset, we first downsample point clouds from 20Hz to 10Hz in order to acquire larger motion between input frames and align with the other two datasets. We select the point clouds at a given interval of frames from the 10Hz point cloud as input, and the remaining point clouds as the ground truth of interpolation. Particularly, the frequency of input point clouds is 2.5Hz when there are three interpolation frames to predict between the middle two input frames.

Our NL-Drive dataset is intended to focus on large movements in as many autonomous driving scenarios as possible. Thus, we try to sort out hard samples that possess the largest relative pose transformation between frames while ensuring it is above the selection threshold from all scenes of the data source. These samples tend to contain nonlinear motions under the precondition of ego-vehicle large motions. The details for constructing NL-Drive dataset are as follows. We take the standard case in the main paper as an example, i.e., a sample contains 4 frames as multi-frame input and 3 frames between the middle two input frames to interpolate. First, we calculate the 6-DOF relative pose transformation between each two input frames. Then, we transform the relative ego-vehicle pose to the LiDAR sensor coordinate system, in which the rotation is uniformly expressed as the Euler angle and the translation as the translation vector, indicating the rotational angular velocity and translational velocity of the ego-vehicle between two frames to some extent. In this form, we can intuitively infer and compare the magnitude of movements from the value. Next, considering that the pitch and roll rotation in autonomous driving datasets is much slighter than the yaw rotation, we define the metric for the rotational motion as the yaw angle. The metric for translational motion is defined as the root-mean-square of the translation vector. We select out the top-$\mathit{k}$ samples with the largest rotational or translational values from each scene and filter them with the threshold ($5.0^{\circ}$ for yaw, $2.5\mathit{m}$ for translation) that is utilized to balance the greater coverage of scenes and the need for large movements. Finally, we exhibit the local zoom-in view of typical samples from NL-Drive dataset in~\cref{fig:sample}, from which the nonlinear motion of objects can be evidently observed.

\section{Method Details}
\label{sec:appendix method details}

In this section, we elaborate on the network architecture of NeuralPCI and more details of other baselines. 

\subsection{Network Architecture of NeuralPCI}

\begin{table}[t]
\renewcommand\arraystretch{1.1}
\centering
\caption{\textbf{Results with different layer widths for NeuralPCI.} }
\label{tab:supp layer width}
\scalebox{0.9}{
\begin{tabular}{p{0.1\textwidth}<{\centering}p{0.05\textwidth}<{\centering}p{0.07\textwidth}<{\centering}p{0.05\textwidth}<{\centering}p{0.07\textwidth}<{\centering}} 
\hline
\multirow{2}{*}{\begin{tabular}[c]{@{}c@{}}layer\\width\end{tabular}} & \multicolumn{2}{c}{DHB ($\times10^{-3}$)} & \multicolumn{2}{c}{NL-Drive}  \\ 
\cline{2-5}
 & CD$\downarrow$ & EMD$\downarrow$ & CD$\downarrow$ & EMD$\downarrow$ \\ 
\hline
128 & 0.563 & 3.931 & 0.828 & 105.230 \\
256 & 0.547 & 3.766 & \textbf{0.768} & 102.173 \\
512 & \textbf{0.541} & 3.677 & 0.801 & \textbf{97.029} \\
1024 & 0.542 & \textbf{3.651} & 0.770 & 97.099 \\
\hline
\end{tabular}}
\end{table}

\begin{table}[t]
\renewcommand\arraystretch{1.1}
\centering
\caption{\textbf{Results with different layer depths  for NeuralPCI.} }
\label{tab:supp layer depth}
\scalebox{0.9}{
\begin{tabular}{p{0.1\textwidth}<{\centering}p{0.05\textwidth}<{\centering}p{0.07\textwidth}<{\centering}p{0.05\textwidth}<{\centering}p{0.07\textwidth}<{\centering}}  
\hline
\multirow{2}{*}{\begin{tabular}[c]{@{}c@{}}layer\\depth\end{tabular}} & \multicolumn{2}{c}{DHB ($\times10^{-3}$)} & \multicolumn{2}{c}{NL-Drive}  \\ 
\cline{2-5}
 & CD$\downarrow$ & EMD$\downarrow$ & CD$\downarrow$ & EMD$\downarrow$ \\ 
\hline
4 & 0.543 & 3.755 & 0.810 & 107.301 \\
6 & \textbf{0.536} & 3.693 & 0.759 & 102.150 \\
8 & 0.541 & \textbf{3.677} & 0.801 & \textbf{97.029} \\
10 & 0.546 & 3.737 & \textbf{0.750} & 97.213 \\
\hline
\end{tabular}}
\end{table}

As shown in the~\cref{fig:network}, NeuralPCI consists of an 8-layer 512-unit MLP, using the LeakyReLU activation function for each layer \textit{except} the last one. Taking one point of the input point cloud as an example, its 3-dimensional spatial coordinate and 1-dimensional temporal coordinate are concatenated and positional encoded (PE) to obtain a 4$\mathit{k}$-dimensional input fed into the MLP. The positional encoding function $\Gamma(\cdot): \mathbb{R} ^n \rightarrow \mathbb{R} ^{nk}$ is shown below:
\begin{equation}
\label{eq: PE}
\begin{split}
    \Gamma(\mathbf{x}) = \left( \mathbf{x} , \sin(\mathbf{x}), \cos(\mathbf{x}), \dots, \sin(2^{m} \mathbf{x}), \cos(2^{m} \mathbf{x}) \right)
\end{split}
\end{equation}
where we set $\mathit{m}$ to 0 for convenience, thus $\mathit{k}$ is equal to 3.

Besides, 1-dimensional interpolation time is concatenated with the 512-dimensional features of the penultimate layer by a skip connection, which is used to regulate the final 3-dimensional motion output. This output is added element-wise to the spatial coordinate of the input point cloud to obtain the final interpolation point. All points in the point cloud are computed in parallel by MLP with \textit{shared weights}, accumulating the gradients and back-propagating to update the parameters of NeuralPCI.

To further investigate the influence of other network structure hyperparameters on the experimental results, we explore the MLP with different layer widths and depths. The results are presented in~\Cref{tab:supp layer width,tab:supp layer depth}, which indicates that the hyperparameters of network structure have a relatively minor impact on the interpolation results. Taking both accuracy and efficiency into account, we choose the structure with the parametric number of 1.847M as depicted in~\cref{fig:network}.

\subsection{Optimization Details}
The computational cost of EMD increases with the number of point clouds, and down-sampling with EMD loss leads to worse results. Therefore, we only use it on DHB dataset (1024 points). In contrast, we utilize smoothness loss on NL-Drive dataset, which adopts k-nearest neighbor (k = 9) to further regulate local rigid motion for autonomous driving scenarios. Empirically, we set the weights $\alpha, \beta, \gamma$ in the total loss to 1, 50, 0 for DHB Dataset and 1, 0, 1 for NL-Drive Dataset during optimization.

\subsection{Other Baselines}
We adopt the linear interpolation results of outstanding scene flow methods NSFP~\cite{li2021neuralSF} and PV-RAFT~\cite{wei2021pv} as baselines, with the consideration of explicit interpolation methods. We calculate both forward scene flow $\boldsymbol{f}_{0 \rightarrow 1}$ and backward scene flow $\boldsymbol{f}_{1 \rightarrow 0}$ from the pair-frame inputs, and interpolate linearly to acquire the scene flow between the reference frames and the interpolation frame, which is used to warp the input frame. Then, as described in \cref{eq:linear intep fwd,eq:linear intep bwd}, we can obtain the intermediate frame based on the forward or backward scene flow.
\begin{equation}
\label{eq:linear intep fwd}
\begin{split}
    \hat{P}_{fwd}=P_{0}+t \times \boldsymbol{f}_{0 \rightarrow 1}
\end{split}
\end{equation}
\vspace{-0.5cm}
\begin{equation}
\label{eq:linear intep bwd}
\begin{split}
    \hat{P}_{bwd}=P_{1}+(1-t) \times \boldsymbol{f}_{1 \rightarrow 0}
\end{split}
\end{equation}
where $t \in (0,1)$ means the time step of the intermediate frame, and $P_0$ and $P_1$ indicate the reference frames before and after the in-between frame.

\section{Supplementary Experiments}
\label{sec:appendix experiments}

In this section, we conduct further supplementary experiments to verify the effectiveness of our method, covering aspects of the convergence, different numbers of input frames, explicit versus implicit frame interpolation, varying point cloud densities and ground point removal.

\subsection{Convergence and Efficiency}

\begin{figure}[t]
\centering
  \includegraphics[width=0.4\textwidth]{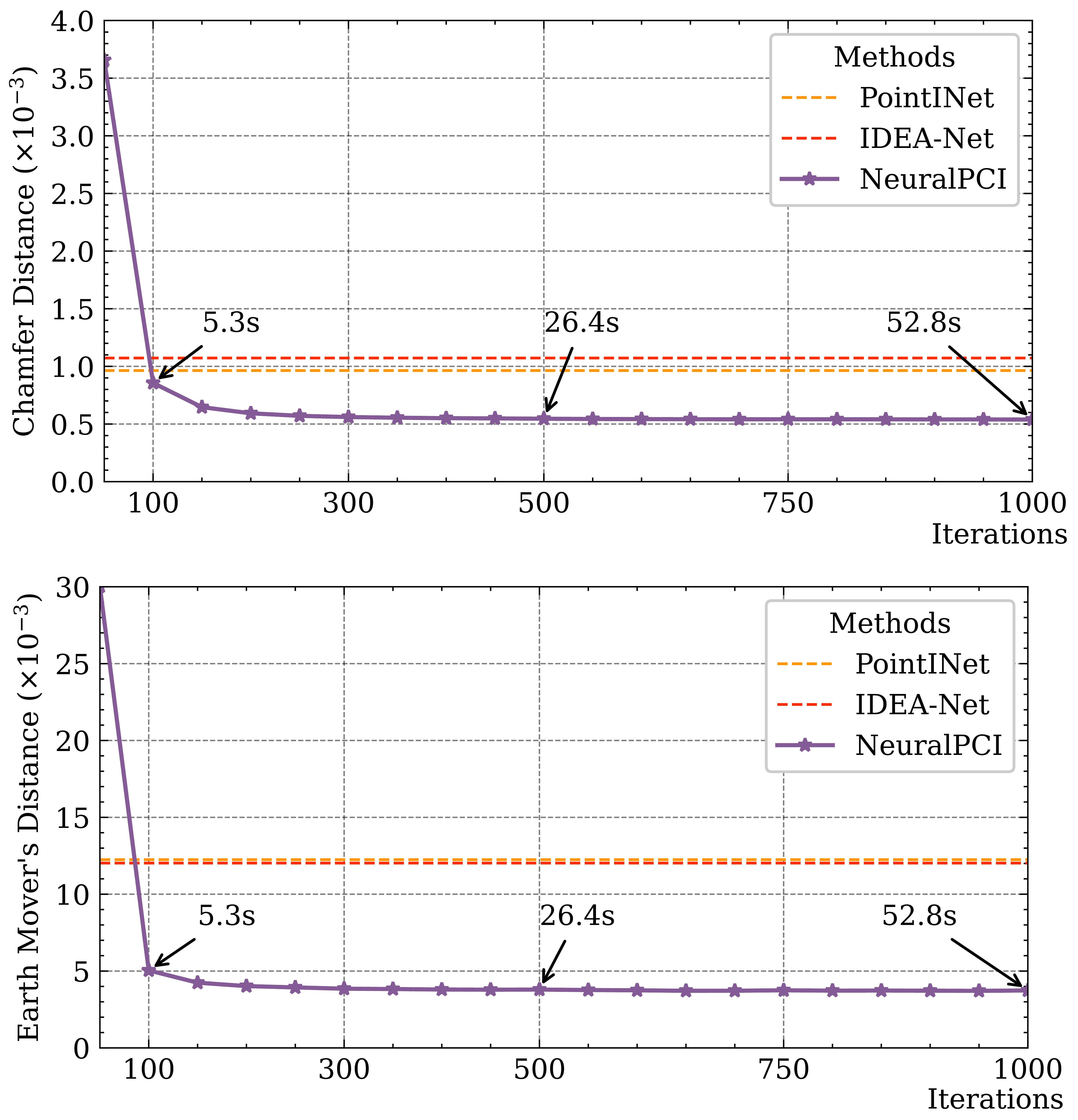}
  \caption{\textbf{Convergence curve of NeuralPCI.} As references, the performance of PointINet~\cite{lu2021pointinet} and IDEA-Net~\cite{zeng2022idea} is denoted by dashed lines. Compared to them, NeuralPCI only needs less than 100 iterations to achieve better CD and EMD results.}
  \label{fig:convergence}
\vspace{-.2cm}
\end{figure}

Since NeuralPCI is optimized at runtime, there is a trade-off between its accuracy and time consumption. We plot the convergence curve of NeuralPCI on DHB dataset in~\cref{fig:convergence}, in which each data point represents the average result of the overall dataset after corresponding iterations. It is evident that NeuralPCI has an excellent convergence, as the error decreases significantly in the first 100 iterations and exceeds previous SOTA methods. When the number of iterations grows, the error gradually reduces, and the convergence is almost complete after 500 iterations. And the entire optimization of 1000 iterations is finished in less than one minute. Therefore, the number of iterations can be determined according to the specific usage scenario. In the case of high timeliness requirements, satisfactory results can be obtained within only 5 seconds, while in other off-board applications, the number of iterations can be appropriately raised to further improve the accuracy.

\subsection{Pair-frame or Multi-frame}

In order to eliminate the unfairness of comparing with the existing pair-frame point cloud interpolation methods, a more comprehensive experiment is conducted. Firstly, using PointINet~\cite{lu2021pointinet} as an example, we input every two frames of all four input point cloud frames to obtain intermediate frame prediction and fuse the interpolation results by random sampling fusion or nearest neighbor (NN) fusion. Random sampling selects points with equal probability from each predicted point cloud result. And NN fusion means to find the nearest point from another point cloud and average the spatial coordinates. Based on this, the pair-frame point cloud interpolation method can also fully utilize the information of all four input point clouds. From the results in~\Cref{tab:NNrandom}, it can be seen that since the final predicted point cloud is located between the second and third input frames, the interpolation results using these two frames as input have the highest accuracy, while those using the first and third frames and using the second and fourth frames have the suboptimal accuracy, and those using the first and fourth frames have the worst accuracy. Secondly, we fuse the aforementioned two or more prediction results by random sampling fusion or nearest neighbor fusion. However, it can be noted that simple point cloud fusion is difficult to achieve higher accuracy.

Finally, we evaluate the results of all methods using both two-frame input and multi-frame input on DHB dataset as well as NL-Drive dataset. As shown in~\Cref{tab:pair-frame or multi-frame}, simply migrating the existing pair-frame interpolation method to multi-frame input does not give better results, while our method can still achieve decent interpolation accuracy even when only using the middle two frames as input. Most importantly, NeuralPCI provides a better way to integrate the spatio-temporal information of multi-frame input point clouds, achieving 10.0\% CD reduction and 12.4\% EMD reduction on DHB dataset and 4.8\% CD reduction and 2.0\% EMD reduction on NL-Drive dataset compared to pair-frame input, respectively.

\begin{table}
\renewcommand\arraystretch{1.2}
\centering
\caption{\textbf{Quantitative results for PointINet~\cite{lu2021pointinet} with different pair-frame inputs and multi-frame fusion methods.} We denote the input frames as frames 1$\sim$4. Different pair frames are used separately as input to predict the same interpolation frame, and the predicted results A$\sim$D are finally fused.}
\label{tab:NNrandom}
\scalebox{0.85}{
\begin{tabular}{ccccccc} 
\hline
\multicolumn{2}{c}{\multirow{2}{*}{Type}}                                                              & \multirow{2}{*}{\begin{tabular}[c]{@{}c@{}}Input\\Frames\end{tabular}} & \multicolumn{2}{c}{DHB ($\times10^{-3}$)} & \multicolumn{2}{c}{NL-Drive}  \\ 
\cline{4-7}
\multicolumn{2}{c}{}                                                                                   & & CD$\downarrow$ & EMD$\downarrow$ & CD$\downarrow$ & EMD$\downarrow$ \\ 
\hline
\multirow{4}{*}{\begin{tabular}[c]{@{}c@{}}pair-\\frame\end{tabular}} 
 & A & frame 2, 3 & \textbf{0.97} & \textbf{12.23} & \textbf{1.06} & \textbf{101.12} \\
 & B & frame 1, 3 & 1.33 & 12.81 & 1.87 & 125.90 \\
 & C & frame 2, 4 & 1.33 & 12.93 & 1.72 & 129.30 \\
 & D & frame 1, 4 & 3.49 & 21.20 & 4.72 & 227.91 \\ 
\hline
\multirow{6}{*}{\begin{tabular}[c]{@{}c@{}}multi-\\frame\end{tabular}} 
 & \multirow{3}{*}{\begin{tabular}[c]{@{}c@{}}random\\fusion\end{tabular}} 
   & B+C & 1.33 & 15.64 & 1.52 & 112.07 \\
 &
   & A+B+C & 1.19 & 15.85 & 1.25 & 105.19 \\
 &
   & A+B+C+D & 1.46 & 17.47 & 1.62 & 118.44 \\ 
\cline{2-7}
 & \multirow{3}{*}{\begin{tabular}[c]{@{}c@{}}NN\\fusion\end{tabular}}
   & B+C & 1.25 & 12.95 & 1.66 & 127.68 \\
 &
   & A+B+C & \underline{1.05} & \underline{12.34} & \underline{1.17} & \underline{105.00} \\
 &
   & A+B+C+D & 1.17 & 12.54 & 1.29 & 108.91 \\
\hline
\end{tabular}}
\end{table}

\begin{table}
\renewcommand\arraystretch{1.1}
\centering
\caption{\textbf{Quantitative results for pair-frame or multi-frame point clouds as input.} The \textit{4-frame} results are based on the NN-fusion of A+B+C inputs described in~\Cref{tab:NNrandom}.}
\label{tab:pair-frame or multi-frame}
\scalebox{0.85}{
\begin{tabular}{cccccc} 
\hline
\multirow{2}{*}{Input}    & \multirow{2}{*}{Methods} & \multicolumn{2}{c}{DHB ($\times 10^{-3}$)} & \multicolumn{2}{c}{NL-Drive}  \\ 
\cline{3-6}
 & & CD$\downarrow$ & EMD$\downarrow$ & CD$\downarrow$ & EMD$\downarrow$ \\ 
\hline
\multirow{5}{*}{2-frame}       
  & NSFP~\cite{li2021neuralSF} & 1.22 & 7.81 & 1.75 & 132.13 \\
  & PV-RAFT~\cite{wei2021pv} & 0.92 & 6.14 & 1.64 & 140.42 \\
  & PointINet~\cite{lu2021pointinet} & 0.96 & 12.25 & 1.06 & 101.12 \\
  & IDEA-Net~\cite{zeng2022idea} & 1.02 & 12.03 & - & - \\
  & Neural PCI & \underline{0.60} & \underline{4.20} & \underline{0.84} & \underline{98.99} \\
\hline

\multirow{5}{*}{4-frame}    
  & NSFP~\cite{li2021neuralSF} & 1.58 & 8.25 & 2.30 & 149.03 \\
  & PV-RAFT~\cite{wei2021pv} & 1.10 & 6.63 & 1.64 & 144.56 \\
  & PointINet~\cite{lu2021pointinet} & 1.05 & 12.34 & 1.17 & 105.00 \\
  & IDEA-Net~\cite{zeng2022idea} & 1.07 & 12.17 & - & - \\
  & Neural PCI & \textbf{0.54} & \textbf{3.68} & \textbf{0.80} & \textbf{97.03} \\

\hline
\end{tabular}}
\end{table}



\begin{table}
\renewcommand\arraystretch{1.1}
\centering
\caption{\textbf{Quantitative results with different numbers of input frames for NeuralPCI.} Among them, \textit{2 frames} input indicates the pair frame setting and \textit{4 frames} input is the standard setting in our main paper. }
\label{tab:input frames}
\scalebox{0.9}{
\begin{tabular}{p{0.12\textwidth}<{\centering}p{0.05\textwidth}<{\centering}p{0.07\textwidth}<{\centering}p{0.05\textwidth}<{\centering}p{0.07\textwidth}<{\centering}}
\hline
\multirow{2}{*}{Input Frames} & \multicolumn{2}{c}{DHB ($\times10^{-3}$)} & \multicolumn{2}{c}{NL-Drive}  \\ 
\cline{2-5}
 & CD$\downarrow$ & EMD$\downarrow$ & CD$\downarrow$ & EMD$\downarrow$ \\ 
\hline
2 frames & 0.60 & 4.20 & 0.84 & 98.99 \\
4 frames & \textbf{0.54} & \textbf{3.68} & \textbf{0.80} & \textbf{97.03} \\
6 frames & 0.55 & 3.74 & 0.86 & 98.82 \\
8  frames & 0.57 & 3.87 & 0.96 & 104.44 \\
\hline
\end{tabular}}
\end{table}


\subsection{More Input Frames}

Beyond the standard four-frame input, our proposed NeuralPCI can also be flexibly extended to more point cloud input frames. Thus, as shown in~\Cref{tab:input frames}, we present the results of NeuralPCI on two datasets with more frames of point clouds as input. Nevertheless, since the predicted interpolation point cloud is always between the two point clouds directly in the middle, the multi-frame inputs that are too far away from it are dramatically less relevant in terms of motion and do not contribute better information to assist interpolation. Besides, the limited modeling capability of MLP causes performance degradation. As a result, the four-frame input is the most appropriate.

\begin{table}
\renewcommand\arraystretch{1.2}
\centering
\caption{\textbf{Quantitative results for explicit and implicit interpolation.} \textit{Ex} indicates explicit interpolation, and \textit{Im} indicates implicit interpolation.}
\label{tab:Explicit interpolation}
\scalebox{0.87}{
\begin{tabular}{cccccc} 
\hline
\multirow{2}{*}{Methods} & \multirow{2}{*}{Type} & \multicolumn{2}{c}{DHB ($\times10^{-3}$)} & \multicolumn{2}{c}{NL-Drive}  \\ 
\cline{3-6}
& & CD$\downarrow$ & EMD$\downarrow$ & CD$\downarrow$ & EMD$\downarrow$ \\ 
\hline
PointINet~\cite{lu2021pointinet} & linear & 0.96 & 12.25 & 1.06 & 101.12 \\
\multirow{3}{*}{Ours (Ex)}    
          & linear & 0.57 & 3.99 & 0.80 & 112.90 \\
          & quadratic & 0.56 & 3.81 & 0.84 & 112.83 \\
          & cubic & 0.60 & 3.90 & 0.89 & 113.24 \\
\multicolumn{1}{c}{Ours (Im)} 
          & neural field & \textbf{0.54} & \textbf{3.68} & \textbf{0.80} & \textbf{97.03} \\
\hline
\end{tabular}}
\end{table}

\begin{table*}[t]
\renewcommand\arraystretch{1.1}
\centering
\caption{\textbf{Quantitative results after removal of ground points on NL-Drive dataset.} }
\label{tab: foreground}
\scalebox{0.95}{
\begin{tabular}{p{0.15\textwidth}<{\centering}p{0.05\textwidth}<{\centering}p{0.08\textwidth}<{\centering}p{0.06\textwidth}<{\centering}p{0.07\textwidth}<{\centering}p{0.06\textwidth}<{\centering}p{0.08\textwidth}<{\centering}p{0.06\textwidth}<{\centering}p{0.07\textwidth}<{\centering}}
\hline
\multirow{2}{*}{Methods} & \multicolumn{2}{c}{Frame - 1} & \multicolumn{2}{c}{Frame - 2} & \multicolumn{2}{c}{Frame - 3} & \multicolumn{2}{c}{Average}  \\ 
\cline{2-9}
 & CD & EMD & CD & EMD & CD & EMD & CD$\downarrow$ & EMD$\downarrow$ \\ 
\hline
PV-RAFT~\cite{wei2021pv} & 1.90 & 150.55 & 2.87 & 217.33 & 2.27 & 253.80 & 2.34 & 207.23 \\
NSFP~\cite{li2021neuralSF} & 1.24 & 137.03 & 2.26 & 198.57 & 3.37 & 256.44 & 2.29 & 197.35 \\
PointINet~\cite{lu2021pointinet} & 1.28 & 138.29 & 1.72 & \textbf{154.32}  & 1.35 & 133.32 & 1.45 & 141.98 \\
\rowcolor{lightgray} NeuralPCI & \textbf{0.92} & \textbf{127.31} & \textbf{1.16} & 167.34 & \textbf{0.91} & \textbf{125.99} & \textbf{1.00} & \textbf{140.21} \\
\hline
\end{tabular}}
\end{table*}

\subsection{Explicit or Implicit Interpolation}

NeuralPCI establishes the motion relationship through an implicit 4D spatio-temporal neural field and derives the corresponding output by varying the interpolation time input. On the contrary, we can also use an explicit approach to model the equations of nonlinear motion and generate the interpolation point clouds at intermediate moments. That is, one of the input point clouds is fed into NeuralPCI as the reference to predict the other three input point clouds, and thus the order of the corresponding points in all the obtained point clouds is aligned with the reference frame. Finally, according to the spatial position of each point at four moments, we employ linear (\cref{eq:explicit linear}), quadratic (\cref{eq:explicit quadratic}) and cubic (\cref{eq:explicit cubic}) equations to describe its motion and calculate the position at the intermediate moment. The equations are as follows:
\begin{equation}
\label{eq:explicit vel}
\begin{split}
    \boldsymbol{v}_{0} = P_{1} - \hat{P}_{0},\quad \boldsymbol{v}_{1} = \hat{P}_{2} - P_{1},\quad \boldsymbol{v}_{2} = \hat{P}_{3} - \hat{P}_{2}
\end{split}
\end{equation}
\vspace{-.7cm}
\begin{equation}
\label{eq:explicit acc}
\begin{split}
    \boldsymbol{a}_{0} = \boldsymbol{v}_{1} - \boldsymbol{v}_{0},\quad \boldsymbol{a}_{1} = \mathbf{v}_{2} - \boldsymbol{v}_{1}
\end{split}
\end{equation}
\vspace{-.7cm}
\begin{equation}
\label{eq:explicit b}
\begin{split}
    \boldsymbol{b} = \boldsymbol{a}_{1} - \boldsymbol{a}_{0}
\end{split}
\end{equation}
\vspace{-.5cm}
\begin{equation}
\label{eq:explicit linear}
\begin{split}
    \hat{P}_{t} = P_{1} + \frac{\left( \boldsymbol{v}_{0} + \boldsymbol{v}_{1} \right) }{2} t
\end{split}
\end{equation}
\vspace{-.5cm}
\begin{equation}
\label{eq:explicit quadratic}
\begin{split}
     \hat{P}_{t} = P_{1} + \frac{\left( \boldsymbol{v}_{0} + \boldsymbol{v}_{1} \right)}{2} t + \frac{\boldsymbol{a}_{0}}{2}  t^{2}
\end{split}
\end{equation}
\vspace{-.5cm}
\begin{equation}
\label{eq:explicit cubic}
\begin{split}
    \hat{P}_{t} = P_{1} + \frac{\left( \boldsymbol{v}_{0} + \boldsymbol{v}_{1} + \boldsymbol{v}_{2} \right)}{3} t + \frac{\left( \boldsymbol{a}_{0} + \boldsymbol{a}_{1} \right)}{4} t^{2} + \frac{\boldsymbol{b}}{6} t^{3}
\end{split}
\end{equation}
where $\hat{P}_{0}, \hat{P}_{2}, \hat{P}_{3}$ are the predicted point clouds of NeuralPCI based on the reference point cloud $P_{1}$, and the points in these four frames located at different time steps posses one-to-one correspondences. Let $t \in (0, 1) $, we calculate the interpolation point cloud $\hat{P}_{t}$ between $P_{1}$ and $\hat{P}_{2}$.

As shown in the~\Cref{tab:Explicit interpolation}, explicit modeling of motion can also yield nice interpolation results, but one single equation can not cover well all the complex motions of the real world, whereas the implicit output of NeuralPCI benefits from the higher order fitting properties of MLP and enables more accurate nonlinear motion estimation for each sample.

\begin{figure}[t]
\centering
  \includegraphics[width=0.4\textwidth]{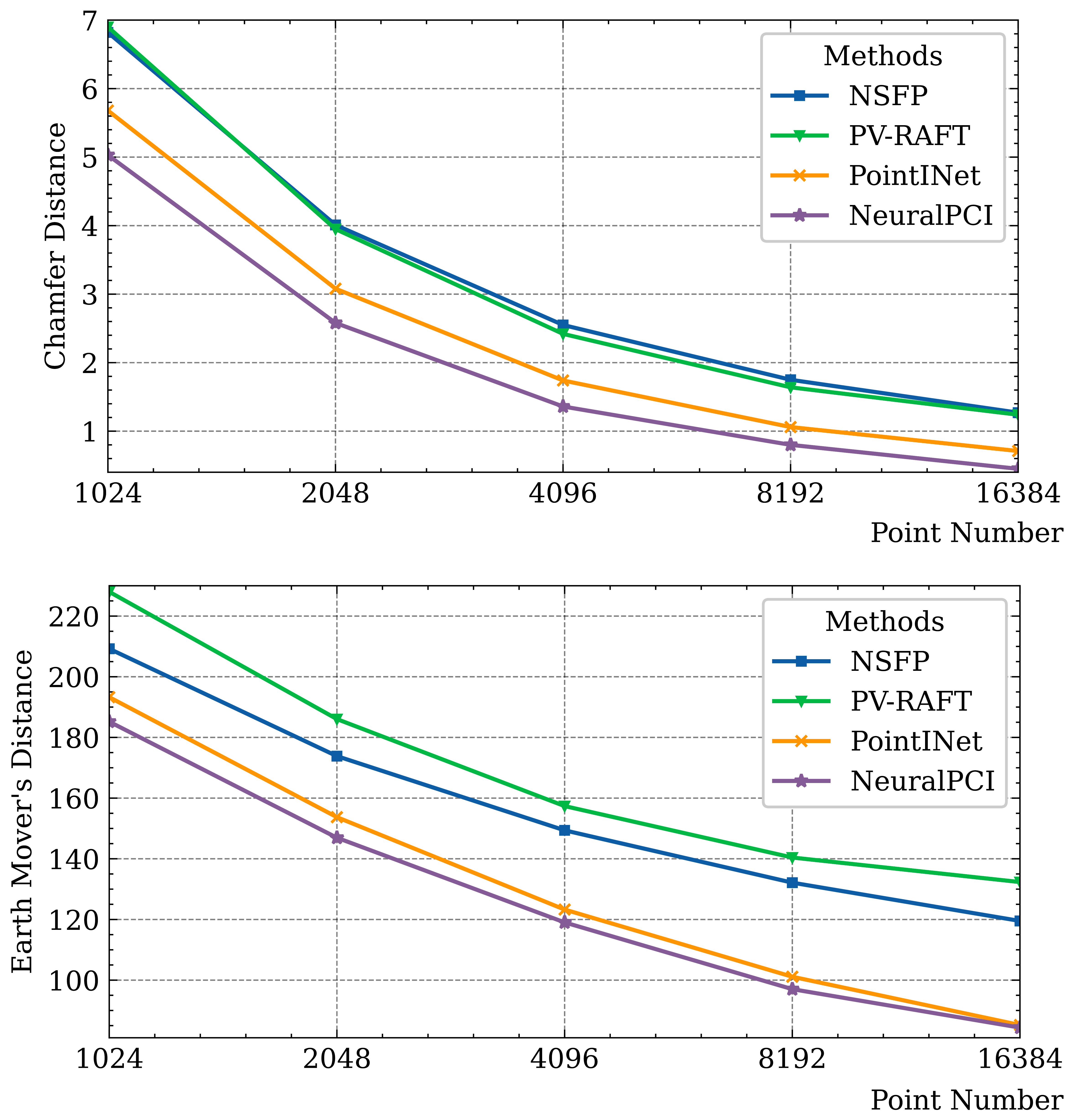}
  \caption{\textbf{Different densities of input point cloud on NL-Drive dataset.} NeuralPCI works and achieves the optimum results with point clouds of diverse densities.}
  \label{fig:point_cloud_density}
\end{figure}

\subsection{Point Cloud Density}

In the comparison experiments of the main paper, each sample of DHB dataset has 1024 points, and the input point cloud of the NL-Drive dataset is sampled uniformly to 8192 points. To further validate the performance of each method under point clouds with different densities, we design a series of experiments with input point clouds of the point number gradients, i.e., 1024, 2048, 4096, 8192, and 16384 points, on NL-Drive dataset for a fair comparison. The results are shown in~\cref{fig:point_cloud_density}, and our proposed NeuralPCI is robust for both sparse and dense point clouds and achieves the state-of-the-art performance.

\subsection{Removal of Ground Points}

Ground points cover a large portion of the point clouds in outdoor autonomous driving scenarios, which remain relatively stationary with respect to the ego vehicle and contain little particular movement information. While it makes sense to recover ground points in the point cloud interpolation, the presence of these static points also affects the demonstration of the advantages of our method for nonlinear motion estimation. Therefore, we remove the ground points from the NL-Drive dataset and conduct the same comparison experiments as shown in the~\Cref{tab: foreground}. The final conclusion remains consistent with the main paper, which shows our method also outperforms previous SOTA methods on dynamic objects after eliminating the influence of static ground points.


\section{Qualitative Results}
\label{sec:appendix qualitative experiments}

\begin{figure*} 
\centering    
\subfloat[] {
\begin{minipage}[b]{0.75\linewidth}
\includegraphics[width=1\columnwidth]{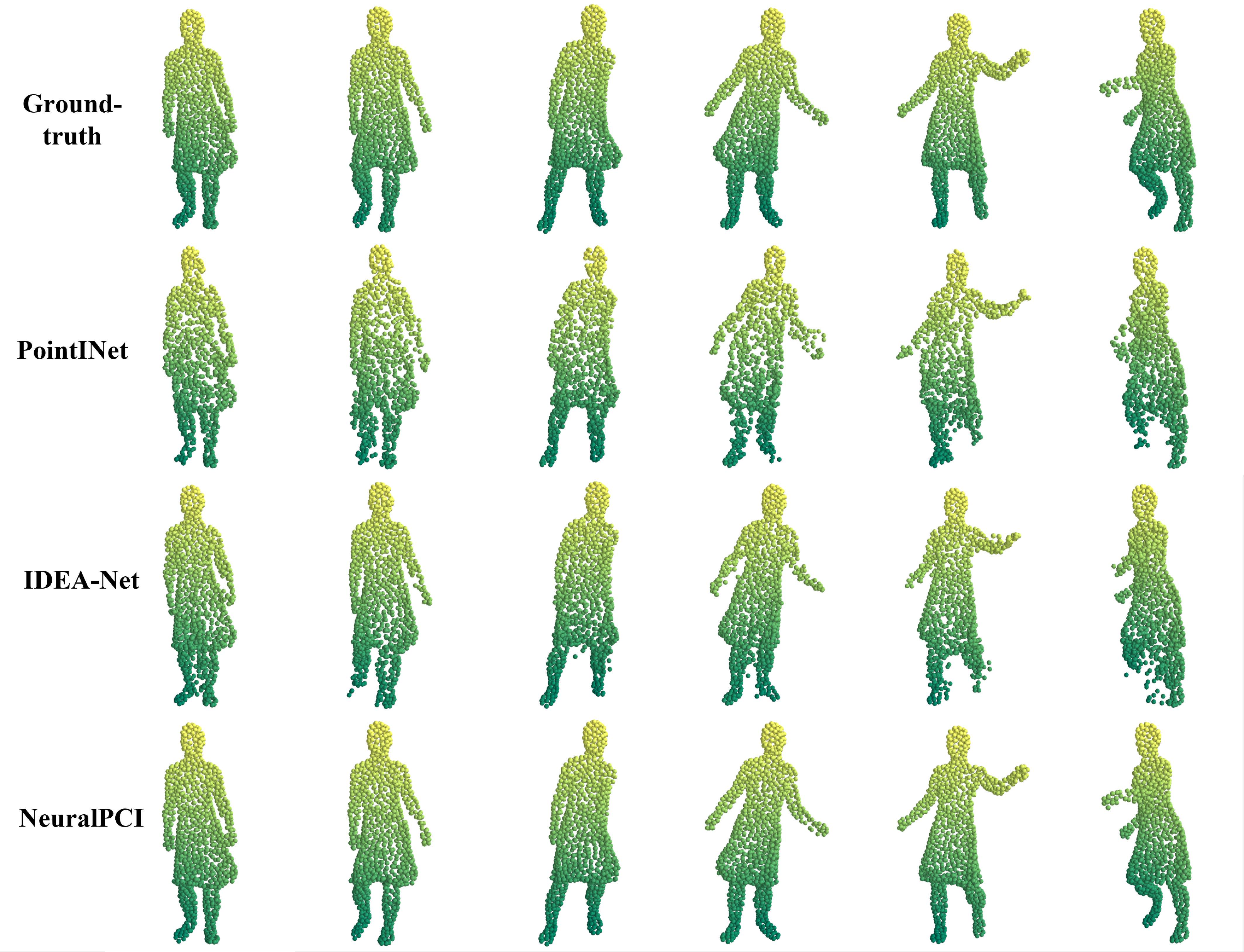}  
\end{minipage}%
}    

\subfloat[] { 
\begin{minipage}[b]{0.75\linewidth}
\includegraphics[width=1\columnwidth]{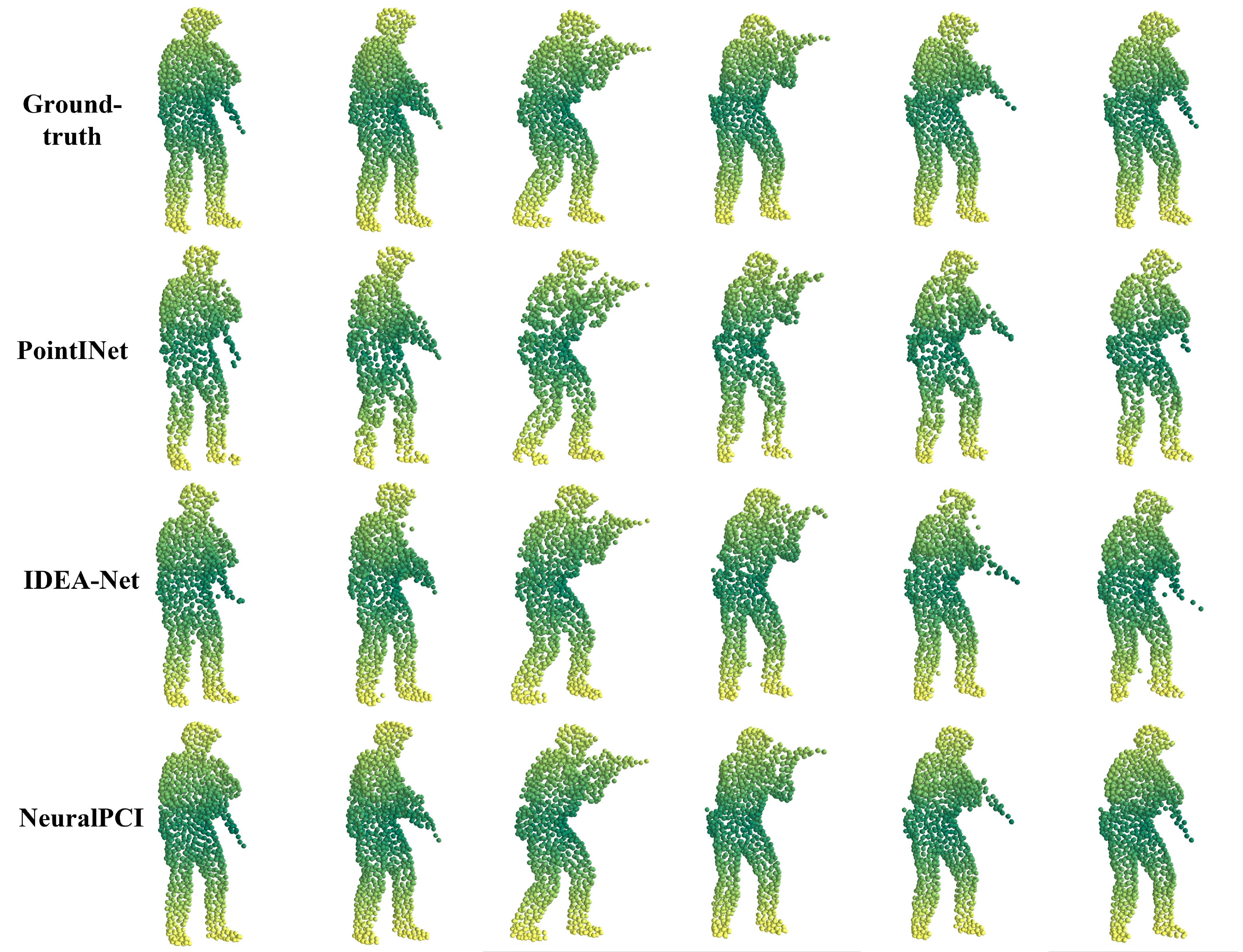} 
\end{minipage}%
}
\caption{\textbf{Qualitative comparison on the test sequence (a) \textit{Swing} and (b) \textit{Soldier} of DHB dataset.} }     
\label{fig:dhb_vis2}  
\end{figure*}

\begin{figure*} 
\centering    
\subfloat[] {
\begin{minipage}[b]{0.8\linewidth}
 \label{fig:a}     
\includegraphics[width=1\columnwidth]{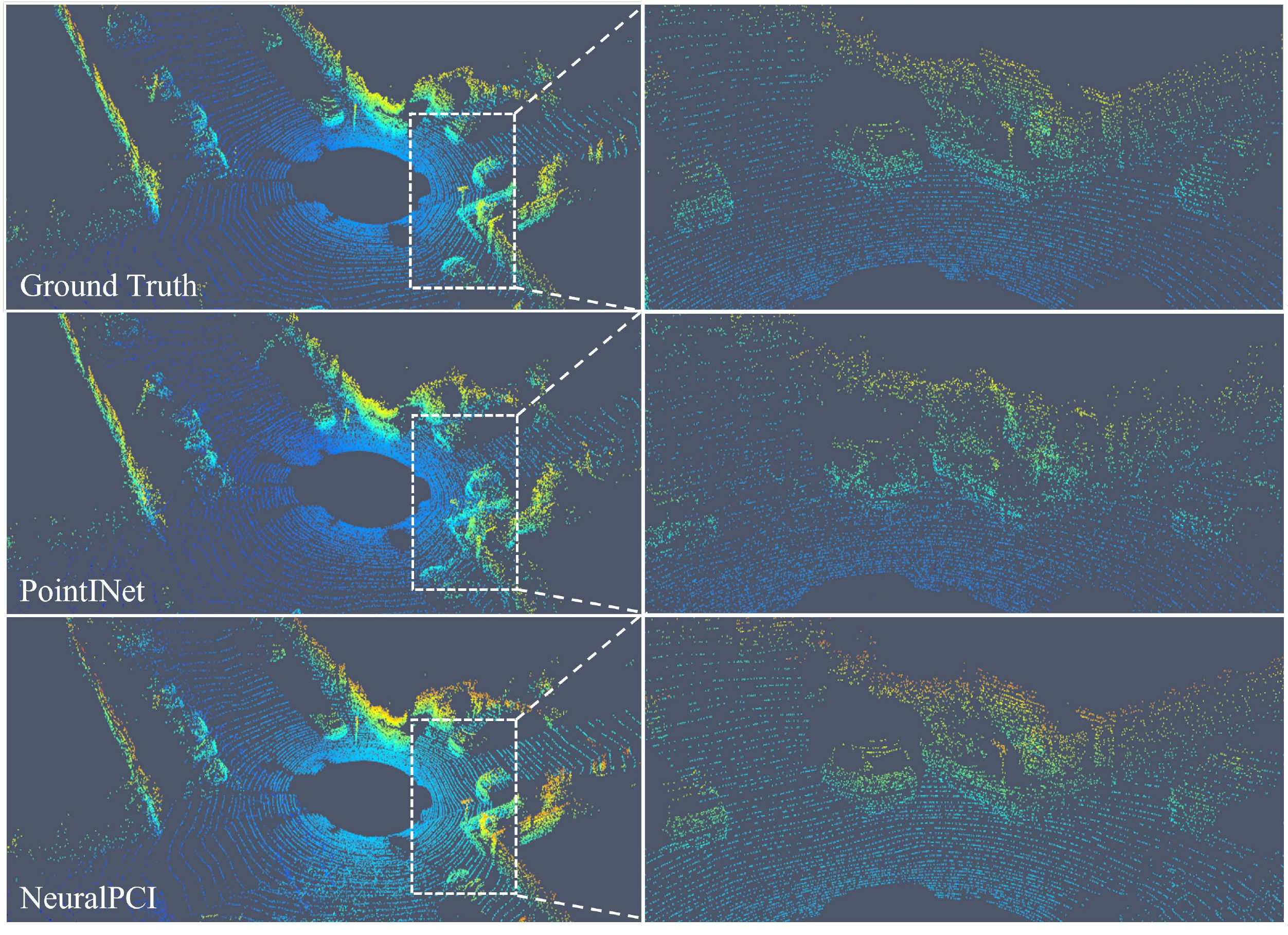}  \end{minipage}%
}    

\subfloat[] { 
\label{fig:b}
\begin{minipage}[b]{0.8\linewidth}
\includegraphics[width=1\columnwidth]{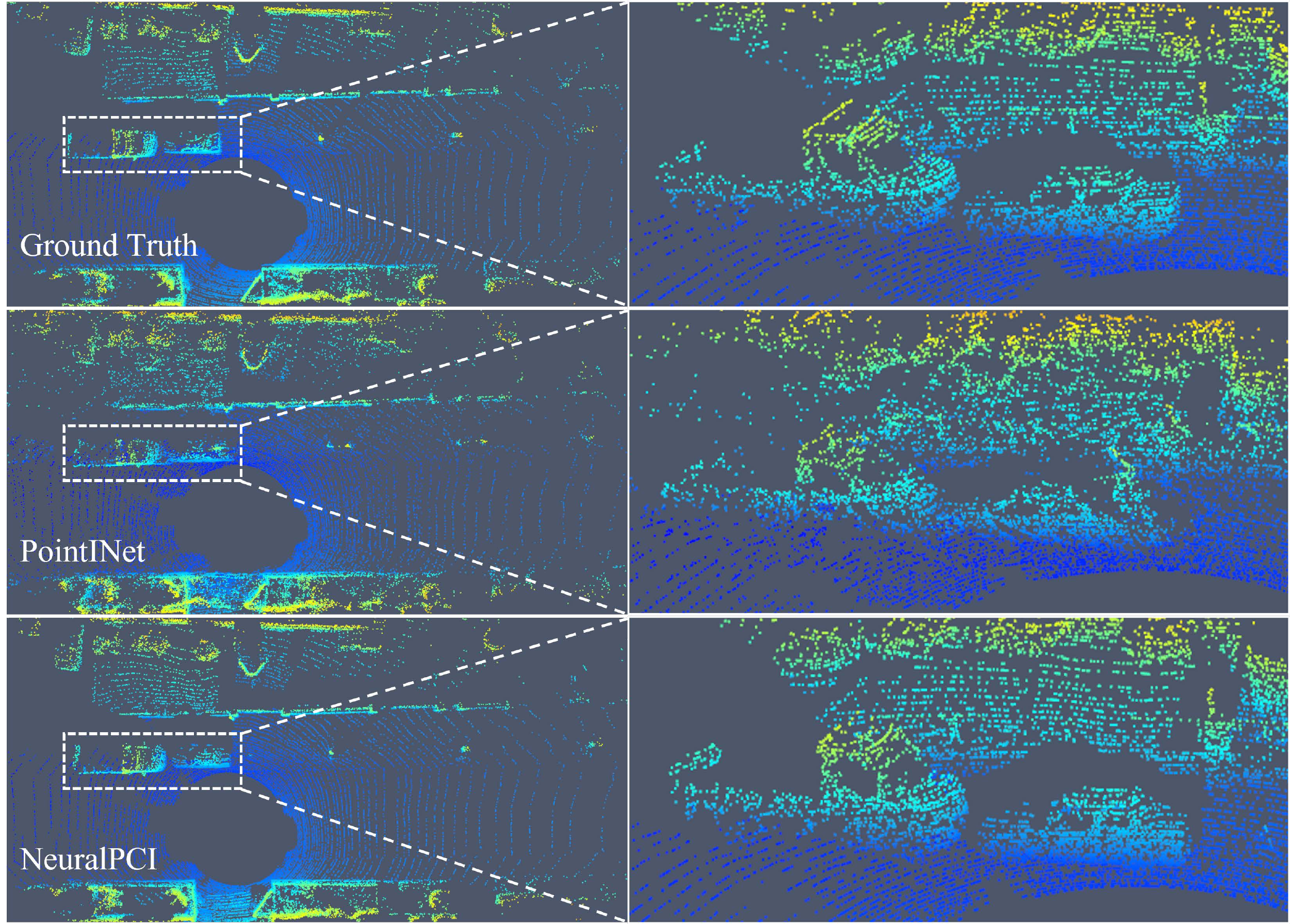} 
\end{minipage}%
}
\caption{\textbf{Qualitative comparison on NL-Drive dataset.} }     
\label{fig:NL-vis}  
\end{figure*}

We provide more qualitative results of our method and baseline methods on DHB and NL-Drive datasets as \cref{fig:dhb_vis2,fig:NL-vis}. It can be seen that the intermediate point cloud frame predicted by NeuralPCI is closest to the ground truth among all the methods on both datasets. On DHB dataset, both the PointINet~\cite{lu2021pointinet} and IDEA-Net~\cite{zeng2022idea} show diverse degrees of outliers (\eg blurry legs and bent guns), especially at the edge of the object, where the motion tends to be larger than the center. On NL-Drive dataset, it is hard for PointINet to predict shape-discernible points, \eg cars in the surroundings, while our method produces fewer distortions and artifacts.

\end{document}